\documentclass[11pt,a4paper]{article}

\usepackage[utf8]{inputenc}
\usepackage[T1]{fontenc}
\usepackage{amsmath,amssymb,amsfonts}
\usepackage{graphicx}
\usepackage{booktabs}
\usepackage{hyperref}
\usepackage{cleveref}
\usepackage{algorithm}
\usepackage{algorithmic}
\usepackage{listings}
\usepackage{xcolor}
\usepackage{tikz}
\usetikzlibrary{arrows.meta,positioning,shapes.geometric,fit,calc}
\usepackage[margin=1in]{geometry}
\usepackage{natbib}
\usepackage{tcolorbox}
\usepackage{enumitem}
\usepackage{multirow}
\usepackage{amsthm}
\newtheorem{definition}{Definition}
\newtheorem{property}{Property}

\newcommand{\faos}{\textsc{FAOS}}
\newcommand{\onto}{\mathcal{O}}

\newcommand{\ctx}{\mathcal{C}}
\newcommand{\skill}{\mathcal{S}}
\newcommand{\role}{\mathcal{R}}
\newcommand{\domain}{\mathcal{D}}
\newcommand{\interact}{\mathcal{I}}

\title{%
  Ontology-Constrained Neural Reasoning in Enterprise Agentic Systems:\\
  A Neurosymbolic Architecture for Domain-Grounded AI Agents%
}

\author{
  Thanh Luong Tuan%
  \thanks{Corresponding author. Email: tluongtuan@my.ggu.edu.
    ORCID: 0009-0000-1199-837X} \\
  \textit{Golden Gate University, San Francisco} \\
  \textit{Foundation AgenticOS (FAOS)}
  \and
  Abhijit Sanyal \\
  \textit{Associate Director, Data, Digital \& IT} \\
  \textit{Novartis Healthcare Pvt.\ Ltd., Hyderabad, India}
}

\date{May 2026 (v3.0.1)}

\begin{document}
\maketitle

\begin{abstract}
Enterprise adoption of Large Language Models (LLMs) is constrained by
hallucination, domain drift, and the inability to enforce regulatory
compliance at the reasoning level. We present a \textbf{neurosymbolic
architecture} implemented within the Foundation AgenticOS (\faos{})
platform that addresses these limitations through
\emph{ontology-constrained neural reasoning}. We introduce a
three-layer ontological framework---Role, Domain, and Interaction
ontologies---grounding LLM-based enterprise agents. We formalize
\emph{asymmetric neurosymbolic coupling}: current enterprise systems
constrain agent inputs (context assembly, tool discovery, governance
thresholds) but not outputs, and we propose mechanisms extending this
coupling to output-side validation (response checking, reasoning
verification, compliance enforcement). A controlled experiment (1,800
runs across five industries and three LLMs: Claude Sonnet~4,
Qwen~2.5~72B, Gemma~4~26B) finds ontology-coupled agents significantly
outperform ungrounded agents on Metric Accuracy ($p < .001$) and Role
Consistency ($p < .001$) across all three models with large effect sizes
(Kendall's $W = .46$--$.64$). Improvements are greatest where LLM
parametric knowledge is weakest---particularly in Vietnam-localized
domains, where ontology lift is $2\times$ that of English domains.
Contributions: (1)~a formal three-layer enterprise ontology model;
(2)~a taxonomy of neurosymbolic coupling patterns;
(3)~ontology-constrained tool discovery via SQL-pushdown scoring;
(4)~a proposed framework for output-side ontological validation;
(5)~empirical evidence for the \emph{inverse parametric knowledge}
effect---ontological grounding value is inversely proportional to LLM
training-data coverage of the domain; (6)~cross-model replication
establishing model-independence; (7)~a production system serving 22
industry verticals with 650+ agents.
\end{abstract}


\section{Introduction}
\label{sec:introduction}

The rapid proliferation of Large Language Models (LLMs) in enterprise
settings has exposed a fundamental tension: LLMs excel at natural
language understanding and generation but lack the formal semantic
grounding required for reliable operation in regulated industries
\citep{ji2023hallucination, huang2023survey}. An insurance underwriter
that hallucinates policy terms, a financial advisor that confuses
Basel III ratios, or a healthcare agent that misapplies HIPAA
guidelines poses a compliance and liability risk, not just a quality
problem. The ``neuro-symbolic debate'' has persisted for decades
\citep{garcez2019neural, hitzler2022neuro}: neural approaches offer
flexibility and generalization; symbolic approaches provide
verifiability and formal guarantees. What is new is the
\emph{enterprise context}---agents must operate within complex
organizational structures, adhere to industry-specific regulations,
and produce outputs that withstand audit scrutiny. We contend the
resolution lies in principled integration: a neurosymbolic
architecture where formal ontologies constrain, guide, and verify
neural reasoning. This paper presents such an architecture,
implemented within the Foundation AgenticOS (\faos{}) platform, and
proposes extensions toward advancing neurosymbolic enterprise AI.

\subsection{Enterprise Grounding}

Enterprise AI agents face a unique variant of the grounding problem
\citep{harnad1990symbol}. Unlike general-purpose assistants, enterprise
agents must:

\begin{enumerate}[label=(\roman*)]
  \item \textbf{Speak the domain language}: Use industry-specific
    terminology correctly (e.g., ``combined ratio'' in insurance,
    ``net revenue retention'' in SaaS, ``average length of stay''
    in healthcare).
  \item \textbf{Reason within regulatory bounds}: Ensure outputs comply
    with applicable regulations (SOX, Basel III/IV, HIPAA, GDPR,
    EU AI Act \citep{euaiact2024}).
  \item \textbf{Follow organizational workflows}: Respect handoff
    protocols, approval chains, and escalation paths defined by the
    organization.
  \item \textbf{Adapt to role-specific perspectives}: A CFO and a
    product manager examining the same data should receive analyses
    framed through their respective decision-making lenses.
\end{enumerate}

Current approaches address these requirements through prompt engineering
\citep{wei2022chain, wang2023selfconsistency}, retrieval-augmented
generation (RAG) \citep{lewis2020retrieval, gao2024retrieval}, and
fine-tuning \citep{hu2022lora}. While effective for individual
dimensions, none provides a \emph{unified formal framework} for
enterprise grounding. Prompt engineering is brittle and unverifiable;
RAG retrieves but does not reason; fine-tuning is expensive and
domain-locked.

\subsection{Contributions}

This paper contributes: (1)~a formal three-layer enterprise ontology
model $\onto = \langle \role, \domain, \interact \rangle$ capturing
role-specific reasoning, domain concepts, and interaction protocols
(\Cref{sec:ontology-framework}); (2)~a taxonomy of neurosymbolic
coupling patterns distinguishing input-side, process-side, and
output-side coupling and characterizing current practice as
predominantly input-side (\Cref{sec:coupling-taxonomy});
(3)~ontology-constrained tool discovery via SQL-pushdown scoring over
domain hierarchies, achieving sub-$100$ms discovery across $600$+
registered skills (\Cref{sec:tool-discovery}); (4)~a proposed
closed-loop framework for output-side ontological validation
(\Cref{sec:closed-loop}); (5)~a 1,800-run controlled experiment across
five regulated industries (including two Vietnamese-language domains)
and three LLM architectures, demonstrating significant improvements on
Metric Accuracy and Role Consistency ($p < .001$) and identifying an
\emph{inverse parametric knowledge} effect where grounding value is
greatest where LLM training coverage is weakest
(\Cref{sec:evaluation}); (6)~cross-model replication on Claude
Sonnet~4, Qwen~2.5 72B, and Gemma~4 26B establishing
model-independence (\Cref{sec:crossmodel}); and (7)~production
evidence from a system implementing input-side coupling across 22
industry verticals, 650+ agents, and 7 bounded contexts
(\Cref{sec:architecture}).

\section{Background and Related Work}
\label{sec:related-work}

\subsection{Neurosymbolic AI}

Neural--symbolic integration has been a persistent research theme
since the connectionist-symbolic debates of the 1980s
\citep{fodor1988connectionism}, catalyzed recently by deep learning
maturation, knowledge representation standards (OWL, RDF, SKOS), and
the emergence of LLMs as general-purpose reasoning engines.
\citet{garcez2019neural} distinguish \emph{symbolic[neural]},
\emph{neural[symbolic]}, and \emph{neural:symbolic} (bidirectional)
integration; our work falls primarily in \emph{neural[symbolic]} with
proposals toward \emph{neural:symbolic}. A systematic review of 167
NeSyAI papers \citep{colelough2025nesyai} confirms the convergence.
\citet{hitzler2022neuro} argue that ontologies provide a natural
formalism for this integration---type hierarchies, property
constraints, and inference rules that can bound neural computation;
we extend the argument to the enterprise domain, where ontologies
must also capture organizational roles and interaction protocols.

\subsection{Knowledge-Grounded Language Models}

Knowledge grounding for LLMs has been pursued through four main
mechanisms. \textbf{Retrieval-Augmented Generation (RAG)} augments LLM
context with retrieved documents \citep{lewis2020retrieval,
gao2024retrieval}, addressing factual grounding but not structural or
regulatory constraints. \textbf{Knowledge Graph Integration}
\citep{pan2024unifying, zhang2022greaselm} provides relational
grounding but typically lacks the formal semantics of ontological
reasoning. \textbf{Ontology-Enhanced Prompting} uses ontological
definitions to structure prompts \citep{babaei2023llms4ol};
\citet{liu2025ontotune} extend this with OntoTune (WWW 2025), aligning
LLMs through in-context learning on SNOMED~CT. \textbf{Constrained
Decoding} restricts output tokens to grammatical or structural
constraints \citep{willard2023efficient}---syntactic rather than
semantic control. Broader surveys of tool-augmented LLMs
\citep{mialon2023augmented} and hybrid architectures
\citep{marcus2020next} argue for neural--symbolic combination;
\citet{kambhampati2024llms} demonstrate that LLMs struggle with
planning without external symbolic scaffolding.

Empirical work has begun to validate these approaches quantitatively:
\citet{agrawal2024can} survey KG-augmentation for hallucination
reduction; \citet{venkatesh2026ontollm} show ontology-grounded LLMs
exhibit reduced digression in industrial conversation;
\citet{dacruz2025ontology} find ontology-guided KG construction
outperforms vector retrieval on specialized domains; and
\citet{sansford2024grapheval} propose a KG-based framework for
triple-level hallucination measurement. Our work synthesizes these
within a unified multi-layer ontological framework and targets
enterprise-specific tasks that test the boundaries of parametric
knowledge.

\subsection{Ontology-Grounded RAG and Hallucination Mitigation}

Recent work demonstrates dramatic improvements from ontological
grounding over baseline RAG: \citet{sharma2025ograg} propose OG-RAG
(EMNLP 2025), achieving $+55\%$ fact recall and $+40\%$ response
correctness across four LLMs via ontology-anchored hypergraph
retrieval; \citet{ali2026clinical} report $98\%$ clinical QA accuracy
(vs.\ $37\%$ for ChatGPT-4) through RDF/OWL knowledge graphs,
reducing hallucination from $63\%$ to $1.7\%$;
\citet{amayuelas2025grounding} ground each reasoning step in KG data
for $+26.5\%$ over chain-of-thought. These results validate ontological
structure over unstructured retrieval, but all three use
\emph{single-layer} domain ontologies. Our three-layer framework
extends this by encoding not just \emph{what} to know (domain), but
\emph{how} to reason (role) and \emph{when} to delegate (interaction).

\subsection{Neurosymbolic Integration Taxonomies}

\citet{yang2025nesyai} (IJCAI) propose a three-way taxonomy:
Symbolic$\to$LLM, LLM$\to$Symbolic, and LLM+Symbolic; \citet{hakim2026nesyagentic}
provide a PRISMA-based survey of 178 neurosymbolic agentic AI papers.
Our work falls primarily in Symbolic$\to$LLM at L1--L3 with proposals
toward LLM+Symbolic at L4--L5. Neither survey distinguishes
\emph{input-side} from \emph{output-side} coupling---a distinction
central to our asymmetric coupling analysis
(\Cref{sec:coupling-taxonomy}).

\subsection{Parametric Knowledge and Contextual Interference}

A growing literature documents the complex interaction between
parametric and injected knowledge. \citet{lin2026resisting}
demonstrate LLMs become ``overly reliant on external information,
suppressing internal parametric knowledge'' (\emph{contextual
interference}); their Knowledgeable-R1 achieves $+22.89\%$ by
explicitly resisting this displacement. \citet{du2025context} show
that even with \emph{perfect retrieval}, performance degrades
$13.9$--$85\%$ as context length increases---context volume imposes a
cost independent of content quality. \citet{tang2025parametric} show
parametric representations operate at deeper feed-forward layers than
token-level context; \citet{zhao2025reconciliation} (NeurIPS spotlight)
demonstrate parametric and contextual knowledge route through
\emph{distinct attention heads}, with aligned knowledge accumulating
across layers---the architectural basis for why injection can displace
rather than supplement parametric knowledge. These findings
collectively ground our Inverse PKE: grounding helps most where
parametric knowledge is weakest, and can actively \emph{harm}
performance where it is strong---a non-obvious prediction our
5-industry experiment confirms.

\subsection{Enterprise Ontologies}

Enterprise ontology engineering has a rich history, from the
Toronto Virtual Enterprise (TOVE) \citep{fox1992enterprise} to the
Financial Industry Business Ontology (FIBO) maintained by the EDM
Council \citep{bennett2013financial}. Upper ontologies such as the
Basic Formal Ontology (BFO) \citep{arp2015building} and DOLCE
\citep{borgo2022dolce} provide foundational categories for domain
ontology alignment.

However, existing enterprise ontologies were designed for data
integration and semantic interoperability, not for constraining AI
agent behavior. A recent industry effort, the Agentic Ontology of Work
(AOW) \citep{skan2026aow}, defines eight entity types (Agents, Skills,
Intents, Contexts, Policies, Memory, Confidence, Outcomes) for
enterprise agent orchestration---the closest industry parallel to our
framework, though AOW focuses on workflow automation rather than
neurosymbolic grounding of LLM reasoning. We extend the enterprise
ontology tradition with three additional layers: \emph{role ontologies}
(how domain actors reason), \emph{interaction ontologies} (how actors
coordinate), and \emph{governance constraints} (what regulatory bounds
apply).

\subsection{Multi-Agent Systems and Coordination}

The multi-agent systems (MAS) literature provides formal frameworks
for agent coordination \citep{guo2024largelanguage}. However,
classical MAS assumes agents with well-defined utility functions and
communication protocols. LLM-based
agents introduce stochastic behavior that challenges these assumptions.
Concurrent work addresses this gap from different angles:
\citet{zhou2025metagent} propose Metagent-P, a neuro-symbolic planning
agent that uses symbolic verification to ensure reasoning correctness
before execution, while \citet{peer2025ata} decouple agent architecture
into offline symbolic knowledge base ingestion and online deterministic
execution (ATA). Our work addresses the gap by using ontological
constraints to bound the stochastic behavior of LLM agents within
formally defined operational envelopes, with a focus on enterprise
multi-industry deployment rather than open-world planning.

\section{Three-Layer Enterprise Ontology Framework}
\label{sec:ontology-framework}

We define an enterprise ontology $\onto$ as a triple:

\begin{equation}
  \onto = \langle \role, \domain, \interact \rangle
\end{equation}

\noindent where $\role$ is the \emph{Role Ontology}, $\domain$ is the
\emph{Domain Ontology}, and $\interact$ is the \emph{Interaction
Ontology}. Each layer serves a distinct function in grounding agent
behavior.

\subsection{Layer 1: Role Ontology ($\role$)}
\label{sec:role-ontology}

The Role Ontology encodes how specific organizational roles think,
decide, and communicate. Formally:

\begin{equation}
  \role = \{r_1, r_2, \ldots, r_n\}
\end{equation}

\noindent where each role $r_i$ is defined as a tuple:

\begin{equation}
  r_i = \langle \text{id}, \delta_i, \mu_i, \sigma_i, \epsilon_i, \alpha_i \rangle
\end{equation}

\begin{itemize}
  \item $\delta_i$: \textbf{Decision patterns} --- a set of reasoning
    strategies (e.g., $\{\text{strategic}, \text{data-driven},
    \text{risk-averse}\}$)
  \item $\mu_i$: \textbf{Metrics focus} --- domain-specific KPIs the
    role prioritizes (e.g., $\{\text{ARR}, \text{NPS},
    \text{churn\_rate}\}$)
  \item $\sigma_i$: \textbf{Communication style} --- output framing
    (e.g., executive, technical, compliance-oriented)
  \item $\epsilon_i$: \textbf{Expertise domains} --- knowledge areas
    the role is authoritative in
  \item $\alpha_i$: \textbf{Approval authority} --- decisions the role
    can authorize
\end{itemize}

\begin{tcolorbox}[title=Example: Role Definition (Product Manager in SaaS),
  colback=blue!5, colframe=blue!40]
\small
\begin{verbatim}
product_manager:
  decision_patterns: [strategic, user-centric, data-driven]
  metrics_focus: [ARR, NPS, feature_adoption, churn_rate]
  communication_style: executive
  expertise_domains: [product_strategy, user_research]
  approval_authority: [feature_releases, roadmap_changes]
\end{verbatim}
\end{tcolorbox}

The Role Ontology enables \emph{perspective-aware reasoning}: the
same query, posed to agents grounded in different roles, produces
analyses framed through the role's decision patterns and metric
priorities---ontologically grounded persona specification with formal
properties:

\begin{property}[Role Consistency]
  For query $q$ and role $r_i$, if $q$ invokes metric-relevant or
  decision-relevant reasoning, the response $a$ must reference metrics
  $m \in \mu_i$ and apply decision patterns $d \in \delta_i$.
\end{property}

\subsection{Layer 2: Domain Ontology ($\domain$)}
\label{sec:domain-ontology}

The Domain Ontology captures industry-specific concepts, their
relationships, and associated metrics. Formally:

\begin{equation}
  \domain = \langle \mathcal{V}, E, M, G \rangle
\end{equation}

\begin{itemize}
  \item $\mathcal{V}$: \textbf{Verticals} --- industry segments
    organized hierarchically (e.g., \texttt{fintech.payments.\allowbreak{}card\_networks})
  \item $E$: \textbf{Entities} --- domain concepts with formal
    definitions and relationships (e.g., ``Annual Recurring Revenue:
    annualized value of recurring revenue, MRR $\times$ 12'')
  \item $M$: \textbf{Metrics} --- quantitative measures with healthy
    ranges and world-class benchmarks
  \item $G$: \textbf{Governance constraints} --- regulatory frameworks
    applicable to each vertical
\end{itemize}

The hierarchical vertical structure enables \emph{domain-scoped
reasoning}: an agent operating in \texttt{fintech.payments} inherits
concepts from the parent \texttt{fintech} domain while accessing
payment-specific terminology. This hierarchy is exploited in tool
discovery (\Cref{sec:tool-discovery}).

\begin{equation}
  \text{ancestors}(v) = \{v' \mid v' \text{ is a prefix of } v
    \text{ in the hierarchy}\}
\end{equation}

\subsection{Layer 3: Interaction Ontology ($\interact$)}
\label{sec:interaction-ontology}

The Interaction Ontology formalizes organizational workflows as typed
handoff patterns between roles,
$\interact = \langle H, A, P \rangle$:
$H$ is the set of \textbf{handoff patterns} (directed edges between
roles with trigger conditions, required artifacts, and approval
flags); $A$ is the set of \textbf{approval chains} (ordered sequences
of roles that must authorize decisions, with timeout constraints);
$P$ is the set of \textbf{escalation paths} (fallback routing when
conditions exceed an agent's authority). Each handoff
$h = \langle r_{\text{from}}, r_{\text{to}}, \tau, \phi, \beta \rangle$
specifies trigger $\tau$, required artifacts $\phi$, and approval
requirement $\beta$---e.g., a design-to-development handoff from
\texttt{ux\_designer} to \texttt{senior\_developer} triggered on
\texttt{design\_complete}, requiring UI designs, design specs, and
flows, with approval by product manager as escalation path.

\subsection{Ontology Composition Across Industries}

A key property is \emph{composability}: the same structural schema
supports radically different industries by varying per-layer content,
e.g.,
$\onto_{\text{fintech}} = \langle \role_{\text{fintech}},
\domain_{\text{fintech}}, \interact_{\text{fintech}} \rangle$ vs.\
$\onto_{\text{healthcare}} = \langle \role_{\text{healthcare}},
\domain_{\text{healthcare}}, \interact_{\text{healthcare}} \rangle$.
The \faos{} platform currently instantiates this framework across 22
industry verticals (banking, healthcare, fintech, manufacturing,
insurance, software, retail, and $15$ others),\footnote{The empirical
evaluation in \S\ref{sec:evaluation} samples $5$ of these (SaaS,
insurance, healthcare, banking\_vn, insurance\_vn) as a balanced
English/Vietnamese cross-section.} demonstrating that the three-layer
schema is \emph{industry-invariant} while the content is
\emph{industry-specific}.

\section{Taxonomy of Neurosymbolic Coupling Patterns}
\label{sec:coupling-taxonomy}

We propose a taxonomy of how symbolic ontological knowledge can be
coupled with neural LLM reasoning in enterprise systems. This taxonomy
distinguishes three coupling points in the agent execution pipeline.

\subsection{Input-Side Coupling (Currently Implemented)}

Input-side coupling constrains what information the LLM receives before
reasoning begins. In \faos{}, three mechanisms implement this.
\textbf{Context injection}: the \texttt{ContextResolver} loads the
industry ontology at runtime, and the \texttt{PromptInjector} serialises
symbolic definitions as priority-ordered (Role~$>$ Domain $>$
Interaction) natural-language context,
\begin{equation}
  \ctx_{\text{injected}} = f_{\text{inject}}(\role_{r_i}, \domain_v,
    \interact_{h}) \quad \text{s.t.} \quad
    |\ctx_{\text{injected}}| \leq T_{\max},
\end{equation}
with token budget $T_{\max}$ (default $2000$).
\textbf{Tool discovery filtering}: ontological domain hierarchies
filter which tools are visible to an agent
(\Cref{sec:tool-discovery}), ensuring agents invoke only capabilities
relevant to their operational domain.
\textbf{Governance thresholds}: domain-specific quality gates prevent
low-quality skills from being available in regulated industries,
\begin{equation}
  \text{available}(s, d) \iff \text{quality}(s) \geq
    \theta_{\text{gov}}(d),
\end{equation}
where regulated domains (fintech, healthcare) have strictly higher
thresholds $\theta_{\text{gov}}(d)$.

\subsection{Process-Side Coupling (Partially Implemented)}

Process-side coupling constrains the reasoning process itself. In
\faos{}, this is partially implemented through:

\begin{itemize}
  \item \textbf{Autonomy gates}: A block-first approval model where
    sensitive operations require human authorization before execution,
    based on a risk classification matrix.
  \item \textbf{Quality judge verification}: A post-reasoning
    verification node in the agent execution graph that assesses output
    quality (though not ontological compliance).
  \item \textbf{Escalation mechanisms}: When agent confidence falls
    below thresholds, the system escalates to a supervisory agent or
    human operator.
\end{itemize}

\subsection{Output-Side Coupling (Proposed)}

Output-side coupling would constrain LLM outputs against ontological
definitions \emph{after} generation. This is the primary research
frontier and is not yet implemented in \faos{}. We formalize the
requirements in \Cref{sec:closed-loop}.

\subsection{Coupling Maturity Model}

We propose a six-level maturity model for neurosymbolic coupling in
enterprise AI systems:

\begin{table}[htbp]
\centering
\caption{Neurosymbolic Coupling Maturity Model}
\label{tab:maturity}
\begin{tabular}{@{}clp{5.5cm}@{}}
\toprule
\textbf{Level} & \textbf{Name} & \textbf{Description} \\
\midrule
L0 & Ungrounded & LLM operates without ontological context \\
L1 & Context-Injected & Ontology provides prompt context (input-side) \\
L2 & Discovery-Constrained & Ontology filters available tools and
  skills \\
L3 & Process-Gated & Ontology enforces approval gates and escalation
  during execution \\
L4 & Output-Validated & Ontology validates and constrains LLM outputs
  post-generation \\
L5 & Closed-Loop & Full bidirectional coupling: ontology constrains
  inputs, guides reasoning, validates outputs, and evolves from agent
  experience \\
\bottomrule
\end{tabular}
\end{table}

The \faos{} platform currently operates at \textbf{L2--L3}, with
mechanisms for L1 (context injection), L2 (tool discovery filtering),
and partial L3 (autonomy gates, quality judge). This paper proposes
the architecture for L4--L5.

The L2--L3 vs.\ L4--L5 gap in Table~\ref{tab:maturity} exposes the
paper's central structural claim: enterprise systems rigorously
constrain LLM inputs (ontological context, tool filtering, governance
thresholds) but do not validate outputs against the same definitions,
so an agent can receive perfect context yet emit constraint-violating
output. We term this \emph{asymmetric neurosymbolic coupling}, and
close the gap in \Cref{sec:closed-loop}.

\section{System Architecture}
\label{sec:architecture}

The Foundation AgenticOS (\faos{}) is a multi-tenant enterprise
platform (Python/FastAPI, LangGraph orchestration, PostgreSQL, Redis,
Qdrant) comprising $300$+ modules organized into seven bounded
contexts with Anti-Corruption Layers between domains following DDD
\citep{evans2004domain}: \textbf{Ontology Engine} (three-layer
resolution, validation, context injection), \textbf{Skill Registry}
(unified registry with semantic discovery), \textbf{Agent
Orchestration} (LangGraph StateGraph with parallel branching and
quality verification), \textbf{Outcome Tracker} (month-partitioned
immutable event store with provenance chains), \textbf{Tenant Manager}
(five-layer isolation: database, cache, events, API, configuration),
\textbf{Context Engine} (priority-ordered assembly with token-budget
management), and \textbf{Governance} (autonomy gates, cost tracking,
approval workflows).

\subsection{Ontology Resolution Pipeline}

The runtime pipeline is specified in Algorithm~\ref{alg:context-resolution}.

\begin{algorithm}
\caption{Ontology-Constrained Context Resolution}
\label{alg:context-resolution}
\begin{algorithmic}[1]
\REQUIRE User request $q$, tenant $t$, role $r$
\ENSURE Ontologically grounded context $\ctx$
\STATE $\onto_t \gets \textsc{LoadOntology}(t.\text{industry\_id})$
\STATE $\onto_t \gets \textsc{MergeCustomizations}(\onto_t, t.\text{overlays})$
\STATE $r_{\text{def}} \gets \textsc{ExtractRole}(\onto_t.\role, r)$
\STATE $d_{\text{ctx}} \gets \textsc{ResolveDomain}(\onto_t.\domain, q)$
\STATE $i_{\text{ctx}} \gets \textsc{ResolveInteractions}(\onto_t.\interact, r)$
\STATE $\ctx_{\text{raw}} \gets \textsc{Serialize}(r_{\text{def}}, d_{\text{ctx}}, i_{\text{ctx}})$
\STATE $\ctx \gets \textsc{Optimize}(\ctx_{\text{raw}}, T_{\max})$
  \COMMENT{Priority truncation: $\role > \domain > \interact$}
\RETURN $\ctx$
\end{algorithmic}
\end{algorithm}

The pipeline is cached at T1 (in-process, $300$s TTL) and T2 (Redis),
ensuring sub-millisecond resolution for repeated tenant-role queries.

\subsection{Agent Execution Graph}

The LangGraph-based 9-node StateGraph chains: (1)~Intent Classifier,
(2)~Router, (3)~Decomposer, (4)~Plan Generator, (5)~Parallel
Executor, (6)~Specialist Executor, (7)~Quality Judge, (8)~Response
Aggregator, (9)~Escalation Handler. Ontological context is injected
at step~6 via the \texttt{PromptInjector}, and governance constraints
are enforced at step~7 and throughout via autonomy gates.

\section{Ontology-Constrained Tool Discovery}
\label{sec:tool-discovery}

A critical neurosymbolic mechanism in \faos{} is the semantic skill
discovery system, which uses ontological domain hierarchies to filter
and rank available tools.

\subsection{Domain-Hierarchical Scoring}

Skills are tagged with domain paths from the Domain Ontology (e.g.,
\texttt{fintech.payments.card\_networks}). Discovery queries specify
a domain context, and the scoring function exploits the hierarchical
structure:

\begin{equation}
  \text{score}(s, q) = \underbrace{w_1 \cdot
    \text{ts\_rank}(s, q)}_{\text{semantic}} +
    \underbrace{w_2 \cdot \text{domain\_match}(s, q)}_{\text{ontological}}
    + \underbrace{w_3 \cdot \text{cap\_match}(s, q)}_{\text{capability}}
    + \underbrace{w_4 \cdot \text{role\_match}(s, q)}_{\text{role}}
\end{equation}

\noindent where the domain match function implements hierarchical
scoring:

\begin{equation}
  \text{domain\_match}(s, q) = \begin{cases}
    1.0 & \text{if } d_s = d_q \quad \text{(exact match)} \\
    0.5 & \text{if } d_s \in \text{ancestors}(d_q) \quad
      \text{(parent match)} \\
    0.0 & \text{otherwise}
  \end{cases}
\end{equation}

This scoring is implemented as SQL expressions (not Python loops),
enabling efficient database-pushdown evaluation across 600+ registered
skills.

\subsection{Governance-Aware Filtering}

Before scoring, skills are filtered by governance constraints:

\begin{equation}
  \skill_{\text{eligible}} = \{s \in \skill \mid
    \text{quality}(s) \geq \max_{d \in \text{domains}(s)}
    \theta_{\text{gov}}(d)\}
\end{equation}

The ``max rule'' ensures that multi-domain skills (e.g., a skill tagged
with both \texttt{fintech} and \texttt{healthcare}) must meet the
strictest applicable threshold. This provides a formal safety guarantee:
regulated domains cannot be served by skills that meet only the
threshold of a less-regulated domain.

\section{Toward Closed-Loop Neurosymbolic Reasoning}
\label{sec:closed-loop}

The current \faos{} architecture implements input-side neurosymbolic
coupling (L2--L3). We propose extensions toward output-side validation
(L4) and closed-loop reasoning (L5), formalized architecturally but not
yet implemented.

\subsection{Output-Side Validation and Constraint Checking}

We propose an \texttt{OntologyValidator} component that checks LLM
outputs against Domain Ontology constraints:

\begin{definition}[Ontological Compliance]
  An agent response $a$ is \emph{ontologically compliant} with domain
  $d \in \domain$ if:
  \begin{enumerate}[label=(\alph*)]
    \item All domain terms in $a$ are defined in $d.E$
      (terminological consistency).
    \item All metric references in $a$ cite values within defined
      ranges in $d.M$ (metric validity).
    \item All workflow references in $a$ follow handoff patterns in
      $\interact.H$ (interaction compliance).
    \item All regulatory claims in $a$ reference frameworks in $d.G$
      (governance alignment).
  \end{enumerate}
\end{definition}

For industries with formal regulatory requirements, lightweight OWL
reasoning can verify output compliance:
\begin{equation}
  \text{valid}(a, \onto) \iff \onto \not\models \neg a
\end{equation}
That is, an output is valid if and only if the ontology does not
entail its negation---requiring translation of relevant Domain Ontology
portions into OWL axioms executed by a description logic reasoner
(e.g., HermiT, ELK). A complementary decoding-time approach:
\citet{luo2025gcr} propose Graph-Constrained Reasoning (ICML 2025),
using a KG-Trie index to constrain LLM decoding to valid knowledge
graph paths. GCR operates at the token level, our proposed validator
at the semantic level---the two are complementary.
\citet{shen2025slot} achieve $99.5\%$ schema accuracy through
model-agnostic post-processing (SLOT), providing syntactic structuring
that could serve as a preprocessing stage for semantic validation.

\subsection{Ontology Evolution from Agent Experience}

A more ambitious L5 extension is \emph{ontology evolution}: agents
encounter concepts outside the current ontology (discovery), the
system proposes extensions from recurring unrecognized concepts
(proposal), domain experts review and approve (validation), and
approved extensions are merged with provenance tracking (integration).
This creates a \emph{learning ontology} that improves with use,
closing the loop between neural experience and symbolic knowledge.
The architectural implication generalizes an emerging neurosymbolic
intuition: hallucinations arise at perception, not generation---so
L1--L3 coupling constrains agent \emph{perception} via grounding,
and L4--L5 extensions would additionally constrain \emph{generation}
via validation.

\section{Empirical Evaluation}
\label{sec:evaluation}

We evaluate the effect of ontological grounding on enterprise AI agent
performance through a controlled within-subject experiment across five
regulated industries spanning both English-language and Vietnamese-language
domains.

\subsection{Evaluation Metrics}

We define four metrics, each targeting a distinct layer of the ontology:

\paragraph{Terminological Fidelity (TF).}
The proportion of domain terms in agent outputs correctly matching
ontology definitions:
\begin{equation}
  \text{TF}(a, \domain) = \frac{|\text{terms}(a) \cap E|}{|\text{terms}(a)|}
\end{equation}

\paragraph{Metric Accuracy (MA).}
Whether metric references cite values within ontologically defined ranges:
\begin{equation}
  \text{MA}(a, \domain) = \frac{|\{m \in \text{metrics}(a) \mid
    m.\text{value} \in M[m.\text{name}].\text{range}\}|}{|\text{metrics}(a)|}
\end{equation}

\paragraph{Regulatory Compliance (RC).}
Correct citation of applicable regulatory frameworks:
\begin{equation}
  \text{RC}(a, \domain) = \frac{|\text{reg\_refs}(a) \cap G|}{|\text{reg\_refs}(a)|}
\end{equation}

\paragraph{Role Consistency (RS).}
Alignment with the role's decision patterns, KPI focus, and
communication style, averaged across three sub-dimensions:
\begin{equation}
  \text{RS}(a, r_i) = \frac{1}{3}\bigl(\text{decision\_match}(a, \delta_i) +
    \text{kpi\_align}(a, r_i) + \text{style\_match}(a, \sigma_i)\bigr)
\end{equation}

\subsection{Experimental Design}

\paragraph{Conditions.}
We compare four grounding conditions corresponding to maturity levels
in our coupling taxonomy (Section~4):

\begin{enumerate}
  \item \textbf{C1 -- Ungrounded (L0):} System prompt only; no domain context.
  \item \textbf{C2 -- RAG-Only:} Unstructured text chunks extracted from
    ontology blueprint documents (8 curated paragraphs per industry,
    ${\sim}2{,}000$ tokens total, injected as flat reference text).
  \item \textbf{C3 -- Ontology-Coupled (L2):} Structured three-layer
    injection (Role + Domain + Interaction) via the \faos{} PromptInjector
    format (${\sim}2{,}800$--$3{,}200$ tokens depending on industry, vs.\
    ${\sim}2{,}000$ for C2, reflecting structural overhead from section
    headers, typed property-value pairs, and metric range definitions).
  \item \textbf{C4 -- Ontology+Process (L3):} C3 plus a post-generation
    quality judge that scores output quality and flags sub-threshold
    responses for escalation.
\end{enumerate}

\paragraph{Task Set.}
50 tasks across five regulated industries---FinTech (BSA-AML),
Insurance (state-regulated), Healthcare (HIPAA/CMS), Vietnamese
Banking (SBV-regulated), and Vietnamese Insurance (MoF-regulated)---
with 10 tasks per industry spanning terminology (3), metric
interpretation (2), regulatory compliance (2), role-based analysis
(2), and cross-cutting (1) categories. Ground truth is derived from
\faos{} ontology blueprints. Vietnamese industries test grounding in a
domain where LLM training is sparse (SBV circulars, MoF decrees,
bilingual terminology); ontology content is authored in English with
embedded Vietnamese terms, and tasks are posed in English, isolating
\emph{conceptual} knowledge from language translation.

\paragraph{Execution.}
Each task ran under all four conditions with three repetitions
($50 \times 4 \times 3 = 600$ runs per model). The primary agent was
Claude Sonnet~4 \citep{anthropic2025claude} (temperature 0.3); for
model-independence, the full experiment was replicated on Qwen~2.5
72B \citep{qwen2025qwen25} and Gemma~4 26B (MoE, 4B active), for
1{,}800 total runs. An independent judge LLM (Claude Sonnet~4,
temperature 0.0) scored all responses against ground truth using
metric-specific rubrics \citep{zheng2023judging}; cross-cutting tasks
were scored on all four metrics. Three design choices mitigate known
LLM-as-judge reliability concerns \citep{schroeder2024judge}:
temperature 0.0 (deterministic), three repetitions, and structured
JSON rubric schemas. The same judge evaluates all conditions,
eliminating inter-judge variance.

\subsection{Results}

\subsubsection{Overall Effect of Ontological Grounding}

\begin{figure}[htbp]
\centering
\includegraphics[width=0.55\textwidth]{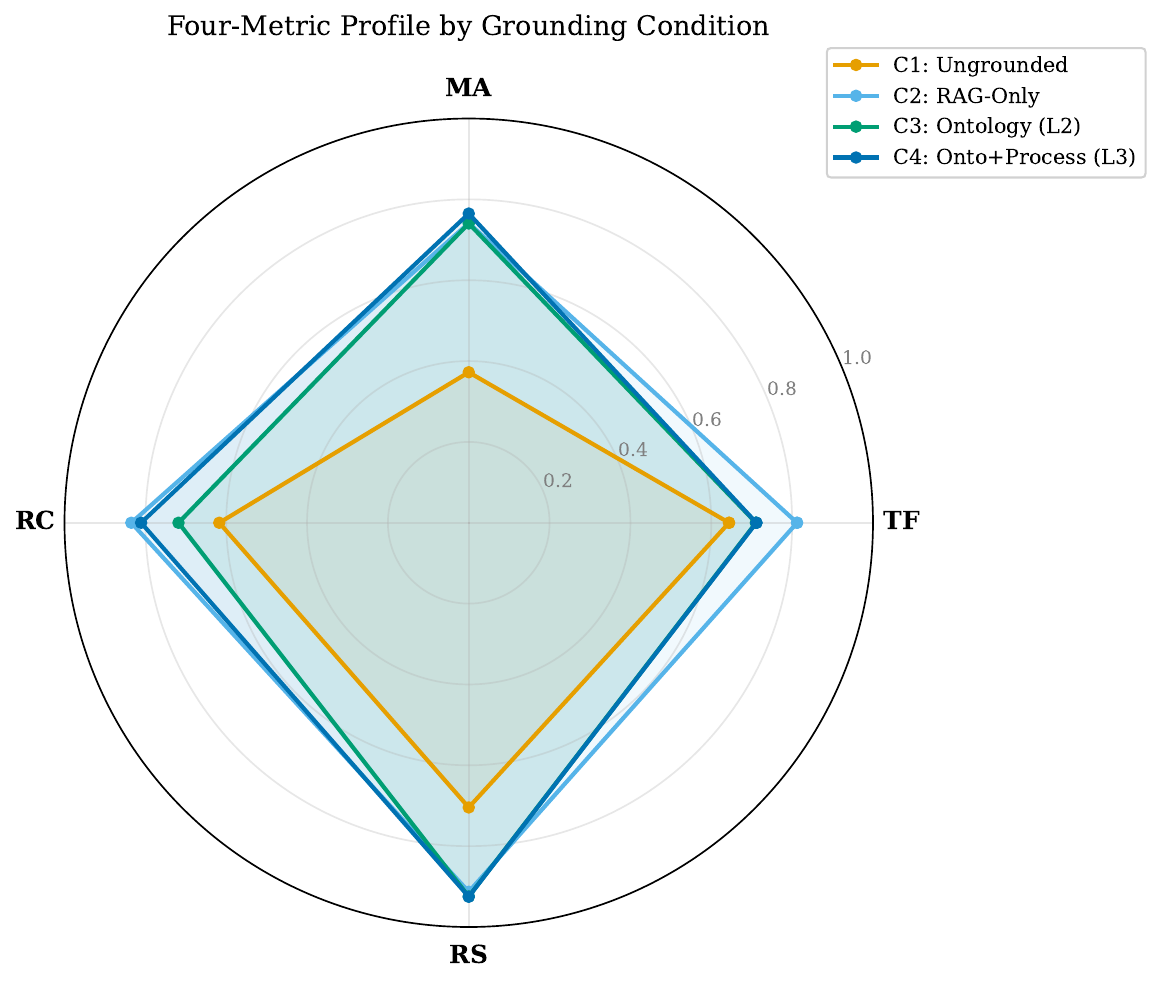}
\caption{Four-metric profile by grounding condition (5 industries, 600 runs, primary model).
C2 (RAG) and C3 (Ontology) both expand the profile relative to C1
(Ungrounded), with the largest gains on MA and RS. C4 (Ontology+Process)
closely tracks C3.}
\label{fig:radar}
\end{figure}

Figure~\ref{fig:radar} visualizes the four-metric profile across
conditions. Table~\ref{tab:friedman} reports Friedman test
\citep{friedman1937use} results: all four metrics achieve omnibus
significance ($\alpha = 0.05$), with MA and RS reaching $p < .001$.

\begin{table}[htbp]
\centering
\caption{Friedman test \citep{friedman1937use} results ($n$ = tasks per metric,
$k = 4$ conditions, 5 industries). $W$ = Kendall's coefficient of concordance
(small: ${<}.06$; medium: $.06$--$.14$; large: ${>}.14$).
Post-hoc: Wilcoxon signed-rank with Bonferroni correction ($m = 3$ selected pairs).
Task counts: TF $n = 20$ (15 terminology + 5 cross-cutting); MA, RC, RS each
$n = 15$ (10 category-specific + 5 cross-cutting scored on all metrics).}
\label{tab:friedman}
\begin{tabular}{@{}lcccccl@{}}
\toprule
\textbf{Metric} & $n$ & $\chi^2(3)$ & $p$ & $W$ &
  \textbf{C1 vs.\ C3} & \textbf{Verdict} \\
\midrule
\textbf{TF} & 20 & \textbf{9.42} & \textbf{.024} & \textbf{.157} &
  $p_{\text{corr}} = 1.0$ & \textbf{Omnibus only}$^\dagger$ \\
\textbf{MA} & 15 & \textbf{20.71} & \textbf{$<$.001} & \textbf{.460} &
  $p_{\text{corr}} < .001$*** & \textbf{Significant} \\
\textbf{RC} & 15 & \textbf{14.30} & \textbf{.003} & \textbf{.318} &
  $p_{\text{corr}} = .32$ & \textbf{Omnibus only}$^\ddagger$ \\
\textbf{RS} & 15 & \textbf{27.64} & \textbf{$<$.001} & \textbf{.614} &
  $p_{\text{corr}} = .003$** & \textbf{Significant} \\
\bottomrule
\multicolumn{7}{l}{\footnotesize $^\dagger$Omnibus significant but
  pairwise C1$\to$C3 not significant on any model (see text).} \\
\multicolumn{7}{l}{\footnotesize $^\ddagger$Pairwise C1$\to$C3 significant
  on Qwen 2.5 72B ($p_{\text{corr}} = .019$); see Table~\ref{tab:crossmodel}.}
\end{tabular}
\end{table}

\begin{figure}[htbp]
\centering
\includegraphics[width=\textwidth]{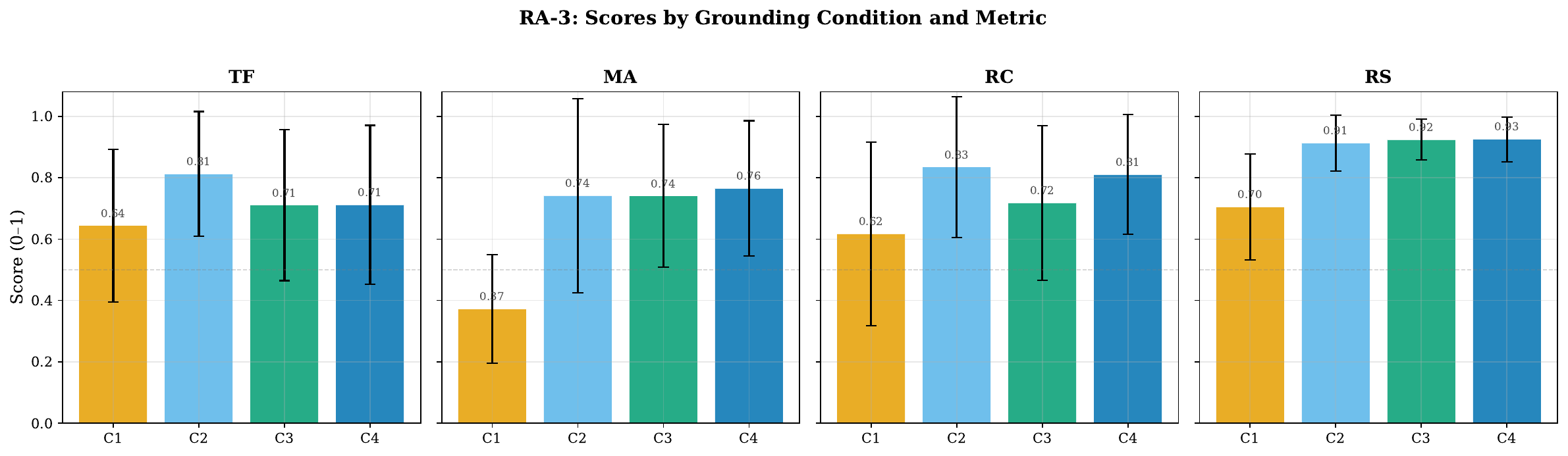}
\caption{Mean scores by condition for each metric (5 industries, 600 runs).
MA and RS show the clearest separation between C1 and grounded conditions.
High variance reflects task-level heterogeneity in LLM parametric
knowledge across English and Vietnamese domains.}
\label{fig:grouped-bar}
\end{figure}

The expanded 5-industry study (600 runs vs.\ the 360-run pilot)
substantially strengthens statistical evidence: $W$ increased for MA
($0.180 \to 0.460$, large) and RS ($0.165 \to 0.614$, large); RC
moved from approaching significance ($p = .072$) to significant
($p = .003$, $W = .318$). TF reached omnibus significance
($p = .024$) but C1$\to$C3 post-hoc failed on all three models
(Claude $p_{\text{corr}} = 1.0$, Qwen $.69$, Gemma $.09$)---consistent
with the Inverse PKE (\Cref{sec:inverse-pke}): for well-established
terminology, parametric knowledge is sufficient. Adding
Vietnamese-language industries, where parametric knowledge is sparse,
amplified effect sizes on the remaining metrics
(Figure~\ref{fig:grouped-bar}).

\subsubsection{Per-Task Analysis}

To understand where ontological grounding provides the greatest value,
Table~\ref{tab:per-task} reports the mean score on each task's
\emph{tested} metric (averaged across 3 repetitions).

\begin{table}[htbp]
\centering
\caption{Selected per-task scores (tested metric, mean of 3 reps).
$\Delta$ = C3 $-$ C1.}
\label{tab:per-task}
\small
\begin{tabular}{@{}llccccc@{}}
\toprule
\textbf{Task} & \textbf{Metric} & \textbf{C1} & \textbf{C2} &
  \textbf{C3} & \textbf{C4} & $\Delta$ \\
\midrule
\multicolumn{7}{l}{\textit{Largest improvements (C3 $>$ C1):}} \\
FT-T10 (BNPL handoff) & MA & .27 & .87 & .93 & 1.00 & \textbf{+.66} \\
IN-T5 (UW cycle time) & MA & .07 & .00 & .69 & .69 & \textbf{+.62} \\
BV-T10 (loan approval) & TF & .30 & .90 & .90 & .90 & \textbf{+.60} \\
HC-T4 (readmission/HAI) & MA & .42 & 1.00 & 1.00 & 1.00 & \textbf{+.58} \\
IVN-T8 (bancassurance) & RS & .39 & .71 & .96 & .91 & \textbf{+.57} \\
\midrule
\multicolumn{7}{l}{\textit{LLM parametric knowledge already strong (C1 $\geq$ C3):}} \\
FT-T7 (PCI-DSS levels) & RC & 1.00 & 1.00 & .42 & .89 & $-.58$ \\
IN-T1 (combined ratio) & TF & .81 & .93 & .50 & .51 & $-.31$ \\
IN-T2 (persistency/lapse) & TF & .87 & 1.00 & .57 & .56 & $-.30$ \\
\bottomrule
\end{tabular}
\end{table}

Two patterns emerge. First, ontological grounding adds the most
value where parametric knowledge is weakest: enterprise-specific
metric benchmarks, organization-specific role decision patterns, and
handoff SLAs---precisely the knowledge types ontologies capture and
training data does not. Second, for well-known concepts (combined
ratio, persistency rate, PCI-DSS) the ungrounded LLM already achieves
high scores and structured injection can \emph{reduce} performance,
likely by displacing parametric knowledge from the effective context
window. This suggests adaptive injection strategies (injecting only
what the LLM is unlikely to know) may outperform blanket injection.

\subsubsection{Industry Comparison}

Table~\ref{tab:industry} reports per-industry mean scores.

\begin{table}[htbp]
\centering
\caption{Mean scores by industry and condition ($n = 120$ per industry,
balanced design). Grand means are simple averages across the five
industries. Each cell averages all four metrics across conditions and
repetitions for that industry.}
\label{tab:industry}
\small
\begin{tabular}{@{}lcccccc@{}}
\toprule
\textbf{Industry} & \textbf{C1} & \textbf{C2} & \textbf{C3} & \textbf{C4} &
  $\Delta_{\text{C1}\to\text{C3}}$ & \textbf{Note} \\
\midrule
FinTech & .66 & .94 & .77 & .85 & +.12 & Strong LLM baseline \\
Insurance & .71 & .78 & .79 & .77 & +.08 & Inverse PKE on TF \\
Healthcare & .63 & .92 & .81 & .84 & +.17 & Large MA gain \\
Banking VN & .47 & .77 & .76 & .81 & +.29 & \textbf{Largest $\Delta$} \\
Insurance VN & .45 & .72 & .74 & .75 & +.28 & \textbf{Strong $\Delta$} \\
\midrule
\textbf{Grand} & \textbf{.58} & \textbf{.83} & \textbf{.77} & \textbf{.80} & \textbf{+.19} & \\
\bottomrule
\end{tabular}
\end{table}

\begin{figure}[htbp]
\centering
\includegraphics[width=\textwidth]{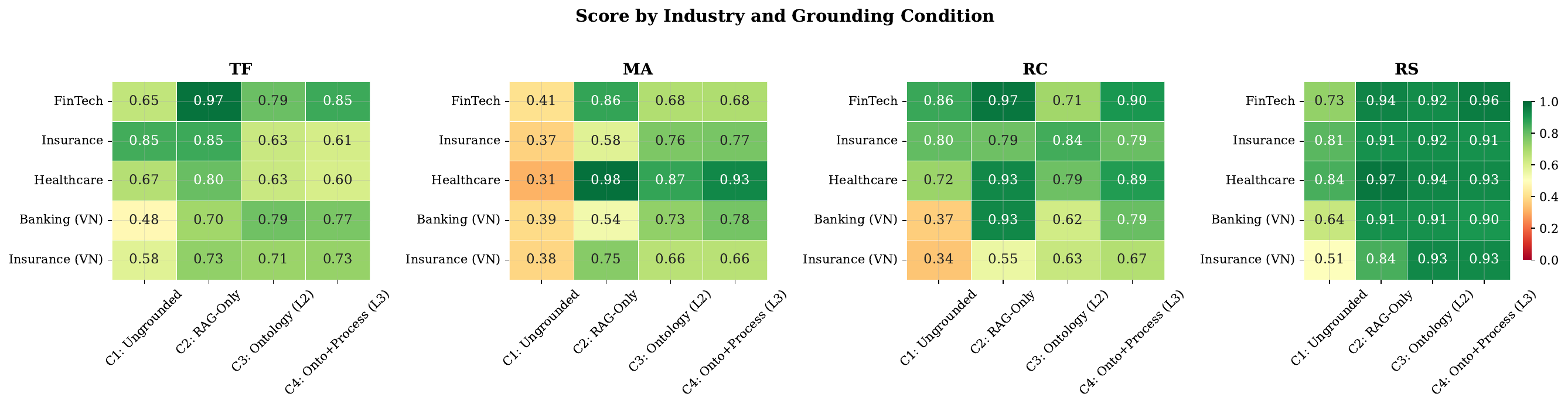}
\caption{C3 (Ontology) scores by industry and metric. Vietnamese industries
(banking\_vn, insurance\_vn) show lower TF scores but strong RS, consistent
with the inverse parametric knowledge effect.}
\label{fig:heatmap}
\end{figure}

\begin{figure}[htbp]
\centering
\includegraphics[width=\textwidth]{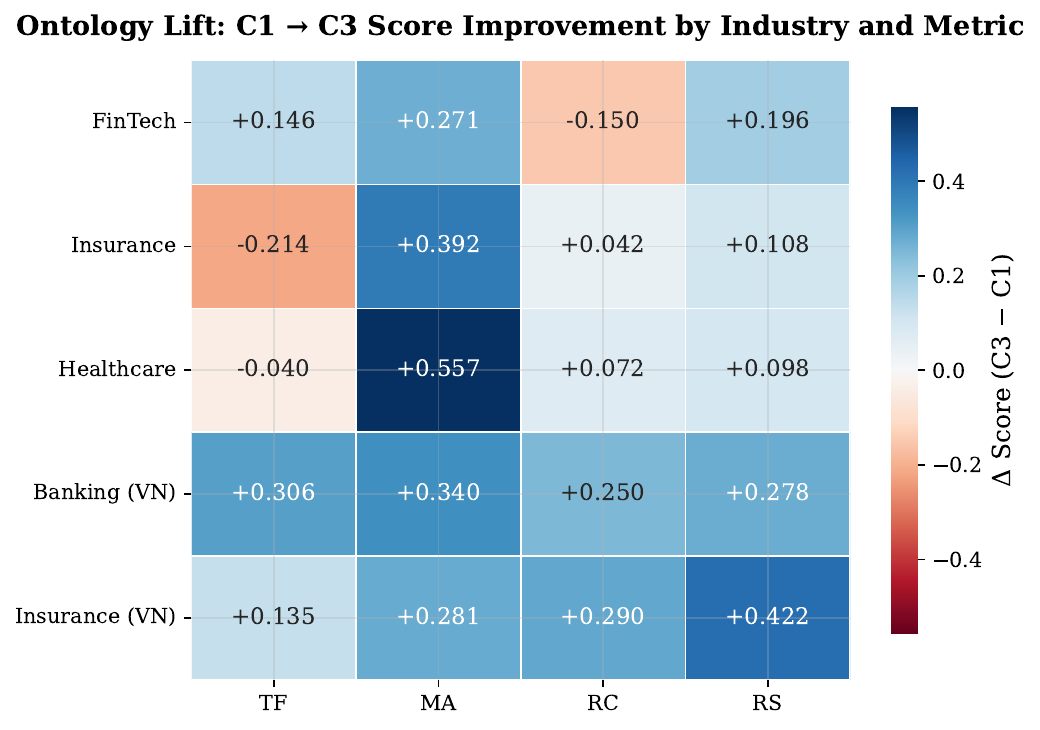}
\caption{Ontology improvement ($\Delta_{\text{C1}\to\text{C3}}$) by industry
and metric. Complements Figure~\ref{fig:heatmap} by visualizing
\emph{change} rather than \emph{absolute} C3 scores: green cells indicate
improvement; red cells indicate regression. Vietnamese industries show
the largest positive deltas on MA and RS, while well-known English
terminology (TF) sometimes regresses, consistent with the Inverse
Parametric Knowledge Effect.}
\label{fig:delta-heatmap}
\end{figure}

Vietnamese industries show the largest improvements: Banking VN
($\Delta = +.29$) and Insurance VN ($\Delta = +.28$) vs.\ the
English-language average ($+.12$)---a $2\times$ amplification
consistent with the Inverse PKE: ontological grounding adds most value
where the pretraining corpus provides least coverage (Vietnamese
regulatory terminology, local metric thresholds, bilingual
role-specific vocabulary). RS improvements are remarkably consistent
across all five industries (C1: $.51$--$.84 \to$ C3:
$.91$--$.94$), suggesting role-specific decision patterns are
uniformly underrepresented in parametric knowledge regardless of
language (Figures~\ref{fig:heatmap}, \ref{fig:delta-heatmap}).

\subsubsection{RAG vs.\ Ontology}

C2 (RAG) is competitive with---and on TF superior to---C3 (Ontology):
on the primary model, C2 TF $= .812$ vs.\ C3 TF $= .711$. Two
mechanisms contribute. First, our RAG chunks were curated from the
same ontology blueprints as C3, making them unusually well-organized
relative to real-world retrieval. Second, structured ontological
format is less token-efficient for terminology recall than
unstructured prose: C3 injects ${\sim}2{,}800$--$3{,}200$ tokens
versus C2's ${\sim}2{,}000$, a $40$--$60\%$ overhead on structural
scaffolding (section headers, property--value pairs, typed
relationships). The C2--C3 post-hoc comparison was not significant on
any metric ($p_{\text{corr}} > 0.20$). A controlled budget-equalized
experiment isolating the format effect from the volume effect remains
future work; we return to the categorical distinction between these
approaches in \Cref{sec:discussion}.

\section{Discussion}
\label{sec:discussion}

\subsection{Why Ontologies, Not Just RAG?}

A natural question is whether formal ontologies add value beyond
well-organized RAG. On the significant metrics (MA, RC, RS), structured
ontology injection (C3) and RAG (C2) achieved comparable scores, and on
the primary model's grand means C2 scores at or above C3 on TF, MA, and
RC; only RS shows a consistent ontology advantage. Our claim is therefore
not that ontologies uniformly outperform RAG at the lexical level, but
that they provide \emph{categorically different} capabilities that a
document-chunk retrieval pipeline cannot:

\begin{enumerate}
  \item \textbf{Structural constraints}: Ontologies define relationships
    (hierarchies, handoff patterns, approval chains) that flat documents
    cannot express. Full governance enforcement requires machine-readable
    ontological definitions.
  \item \textbf{Composability}: The three-layer framework allows
    independent evolution of roles, domains, and interactions. Adding a
    new role does not require rewriting domain documentation.
  \item \textbf{Verifiability}: Ontological definitions are
    machine-readable and can be checked programmatically. RAG documents
    are opaque to formal verification---critical at coupling levels
    L4--L5, where output-side validation requires structured definitions.
\end{enumerate}

Cross-model replication provides empirical support: on Gemma~4, the
C2$\to$C3 comparison for Role Consistency was significant
($p_{\text{corr}} = .006$)---the only metric$\times$model combination
where ontology achieved pairwise significance over RAG. This suggests
the structural value of ontologies manifests most clearly on metrics
requiring relational reasoning rather than lexical recall. Our
findings participate in a broader convergence: \citet{ali2026clinical}
report 98\% clinical QA accuracy via RDF/OWL knowledge graphs,
\citet{amayuelas2025grounding} improve reasoning by $+26.5\%$ through
step-level KG grounding, and \citet{gaurav2025governance} propose a
Governance-as-a-Service framework complementary to our ontology-driven
approach. Table~\ref{tab:grounding-comparison} contrasts RAG,
fine-tuning, and ontological grounding on enterprise-deployment
dimensions.

\begin{table}[htbp]
\centering
\caption{Comparison of Grounding Approaches for Enterprise AI Agents}
\label{tab:grounding-comparison}
\begin{tabular}{@{}lccc@{}}
\toprule
\textbf{Capability} & \textbf{RAG} & \textbf{Fine-tuning} &
  \textbf{Ontology} \\
\midrule
Domain terminology & Partial & Strong & Strong \\
Regulatory constraints & Partial$^\dagger$ & Implicit & Explicit \\
Role-aware reasoning & Indirect$^\dagger$ & None & Native \\
Workflow enforcement & None & None & Native \\
Multi-industry reuse & Low & None & High \\
Formal verifiability & None & None & Yes \\
Maintenance cost & Medium & High & Medium \\
\bottomrule
\multicolumn{4}{l}{\footnotesize $^\dagger$When RAG sources contain
  regulatory content; zero otherwise.}
\end{tabular}
\end{table}

\subsection{The Inverse Parametric Knowledge Effect}
\label{sec:inverse-pke}

Our pilot study reveals a phenomenon we term the \emph{inverse parametric
knowledge effect}: the value of ontological grounding is inversely
proportional to the LLM's pre-existing parametric knowledge of the
domain concept. Four observations support this:

\begin{enumerate}
  \item \textbf{Metric Accuracy benefits most} ($p < .001$, $W = .460$).
    KPI benchmarks (healthy ranges, world-class targets, SLA thresholds)
    are enterprise-specific and rarely appear in public training corpora.
    Symmetrically, Regulatory Compliance shows the \emph{smallest} lift
    among the significant metrics (Claude $W = .318$) because regulatory
    frameworks (Basel~III, HIPAA, PCI-DSS, SBV circulars) are public
    documents widely represented in pretraining---an asymmetry that is
    itself a prediction of Inverse PKE.

  \item \textbf{TF regression on well-known concepts.} For insurance
    terms like ``combined ratio'' and ``persistency rate,'' ontology
    injection \emph{reduced} scores (IN-T1: $.81 \to .50$; IN-T2:
    $.87 \to .57$). We hypothesize a \emph{context displacement}
    mechanism: injected ontological context occupies system prompt
    capacity that would otherwise be available for parametric recall.

  \item \textbf{Vietnamese industries show the largest $\Delta$.}
    banking\_vn ($\Delta = +.29$) and insurance\_vn ($\Delta = +.28$)
    show 2$\times$ the improvement of English-language industries
    (avg $\Delta = +.12$). Vietnamese regulatory concepts (SBV Circular
    11/2021 NPL thresholds, MoF solvency margin per TT 132/2023,
    bancassurance cooling-off periods) are severely underrepresented in
    English-centric pretraining corpora. This finding is supported by
    \citet{xu2024knowledge} (context-memory conflict taxonomy),
    \citet{lin2026resisting} ($+22.89\%$ from resisting contextual
    interference), \citet{du2025context} (13.9--85\% degradation from
    context length), and \citet{augenstein2026interplay} (dynamicity
    predicts parametric-vs-context override).

  \item \textbf{RS improvements are language-invariant.} Role Consistency
    improved by $+.27$ (C1$\to$C3) across all five industries regardless
    of language, with C3 scores in a narrow range. This is consistent
    with role-specific decision patterns being uniformly underrepresented
    in parametric knowledge, though judge-ceiling effects cannot be fully
    ruled out without human validation.
\end{enumerate}

This effect has practical implications: a production injection pipeline
should estimate parametric confidence per concept and suppress injection
for well-known domains, reserving context budget for enterprise-specific
knowledge. For non-English enterprise domains where parametric coverage
is structurally insufficient, ontological grounding becomes part of the
operational requirement rather than an optional enhancement.

\textbf{Toward a formal model.} The interaction between parametric and
injected knowledge resembles wave interference---\emph{constructive} in
low-coverage domains and \emph{destructive} in high-coverage domains.
Our companion work \citep{luong2026context} develops an
information-theoretic model $V^*(\kappa)$ predicting optimal injection
volume as a function of parametric knowledge density; a planned
standalone study \citep{luong2026quantum} formalizes the interference
analogy through a quantum-inspired framework.

\begin{figure}[htbp]
\centering
\includegraphics[width=\textwidth]{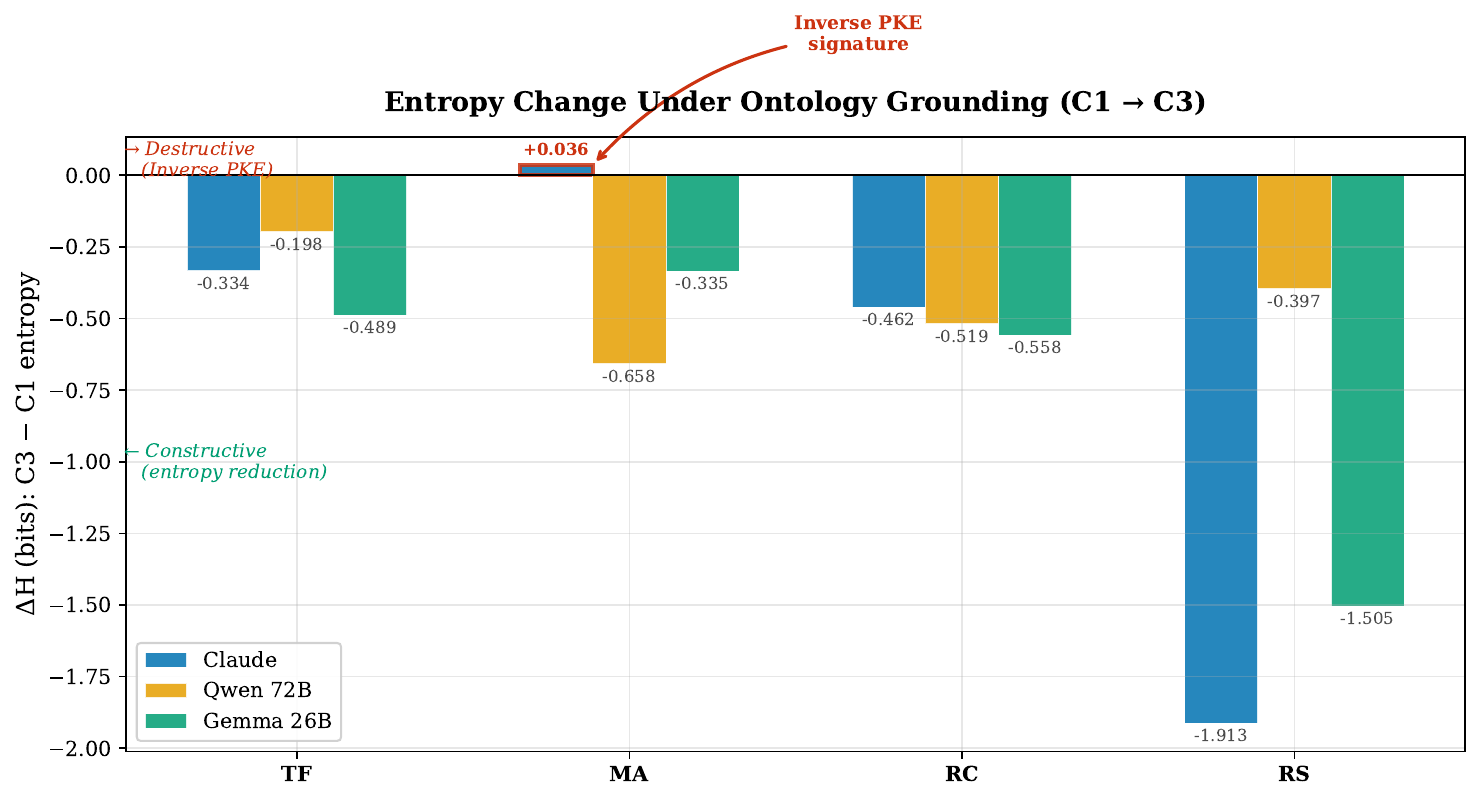}
\caption{Semantic entropy change ($\Delta H$, C1$\to$C3) by metric and
model. Negative values indicate entropy reduction (constructive
grounding); positive values indicate entropy increase (destructive
interference). 11 of 12 metric$\times$model combinations show entropy
reduction. The single exception---MA on Claude---is an empirical
signature of the Inverse PKE: Claude's strong parametric metric knowledge
is disrupted by ontological injection.}
\label{fig:entropy-delta}
\end{figure}

\textbf{Entropy signature.} Following \citet{farquhar2024semantic}, we
computed score distribution entropy $H$ across conditions. Ontological
grounding reduces entropy for 11 of 12 metric$\times$model combinations
($\Delta H < 0$, constructive interference;
Figure~\ref{fig:entropy-delta}). The sole exception is Metric Accuracy
on Claude ($\Delta H = +0.04$ bits)---precisely where Claude's strong
parametric knowledge of industry benchmarks creates destructive
interference. Under a binomial null, 11 of 12 constructive outcomes is
significant ($p = .003$, one-tailed). This \emph{entropy reversal}
constitutes the first empirical entropy signature of Inverse PKE.
Energy-based-model framings of autoregressive LMs
\citep{blondel2025arms} and thermodynamic analyses of LLM training
\citep{liu2025neural} suggest entropy dynamics may generalize as a lens
for Inverse PKE. \citet{wei2025senator} show that structural entropy
along knowledge graph paths identifies LLM knowledge gaps; single-shot
semantic-entropy probes \citep{kossen2025sep} suggest Inverse PKE
detection could operate at inference time.

\subsection{Why C3 and C4 Are Indistinguishable}

A notable non-result is the absence of significant C3$\to$C4 improvement
on any metric ($p_{\text{corr}} > .07$). We attribute this to experimental
design: the C4 quality judge scored outputs and flagged sub-threshold
responses for escalation, but did not alter the response itself---by
design, for fair within-subject comparison. In production deployment,
escalation routes low-quality interactions to supervisory agents or human
operators, producing a qualitatively different user experience not
captured by single-turn scoring. Future evaluation should measure C4's
value through user-facing metrics (escalation rate, downstream task
success).

\subsection{Cross-Model Replication}
\label{sec:crossmodel}

To address model-dependence, we replicated the full 600-run experiment
on two additional open-source models: \textbf{Qwen~2.5 72B}
\citep{qwen2025qwen25} via OpenRouter and \textbf{Gemma~4 26B} (MoE,
4B active) via Google AI Studio. The judge remained fixed on Claude
Sonnet~4 throughout all three replications. Table~\ref{tab:crossmodel}
reports the Friedman test results:

\begin{table}[htbp]
\centering
\caption{Cross-model replication of Friedman test results. Columns
report Friedman $\chi^2(3)$, $p$-value, Kendall's $W$, and post-hoc
C1~vs.~C3 Bonferroni-corrected $p$-value. All three models show
significant omnibus effects on MA and RS; the Inverse PKE replicates
across all models.}
\label{tab:crossmodel}
\small
\begin{tabular}{@{}llcccc@{}}
\toprule
\textbf{Metric} & \textbf{Model} & $\chi^2(3)$ & $p$ & $W$ &
  \textbf{C1 vs.\ C3 ($p_{\text{corr}}$)} \\
\midrule
\multirow{3}{*}{\textbf{TF}}
  & Claude Sonnet 4 & 9.42 & .024* & .157 & 1.000 \\
  & Qwen 2.5 72B    & 13.54 & .004** & .226 & .693 \\
  & Gemma 4 26B     & 10.41 & .015* & .173 & .088 \\
\midrule
\multirow{3}{*}{\textbf{MA}}
  & Claude Sonnet 4 & 20.71 & $<$.001*** & .460 & $<$.001*** \\
  & Qwen 2.5 72B    & 22.52 & $<$.001*** & .501 & $<$.001*** \\
  & Gemma 4 26B     & 25.14 & $<$.001*** & .559 & $<$.001*** \\
\midrule
\multirow{3}{*}{\textbf{RC}}
  & Claude Sonnet 4 & 14.30 & .003** & .318 & .324 \\
  & Qwen 2.5 72B    & 20.07 & $<$.001*** & .446 & .019* \\
  & Gemma 4 26B     & 7.76 & .051 & .172 & .083 \\
\midrule
\multirow{3}{*}{\textbf{RS}}
  & Claude Sonnet 4 & 27.64 & $<$.001*** & .614 & .003** \\
  & Qwen 2.5 72B    & 21.82 & $<$.001*** & .485 & .006** \\
  & Gemma 4 26B     & 28.86 & $<$.001*** & .641 & $<$.001*** \\
\bottomrule
\end{tabular}
\end{table}

Three findings emerge:

\begin{enumerate}
  \item \textbf{MA and RS replicate universally} across all three models
    with highly significant omnibus effects ($p < .001$) and significant
    C1$\to$C3 post-hoc improvements, supporting model-independent value.
  \item \textbf{TF shows omnibus-only significance} across all models
    but no post-hoc C1$\to$C3 gain---consistent with Inverse PKE:
    well-known terminology is already parametrically encoded regardless
    of architecture.
  \item \textbf{RC is model-sensitive.} Claude achieves omnibus
    ($p = .003$) but not post-hoc; Qwen shows both
    ($p_{\text{corr}} = .019$); Gemma approaches significance
    ($p = .051$)---regulatory reasoning benefits depend on the model's
    pre-existing regulatory knowledge.
\end{enumerate}

\begin{figure}[htbp]
\centering
\includegraphics[width=\textwidth]{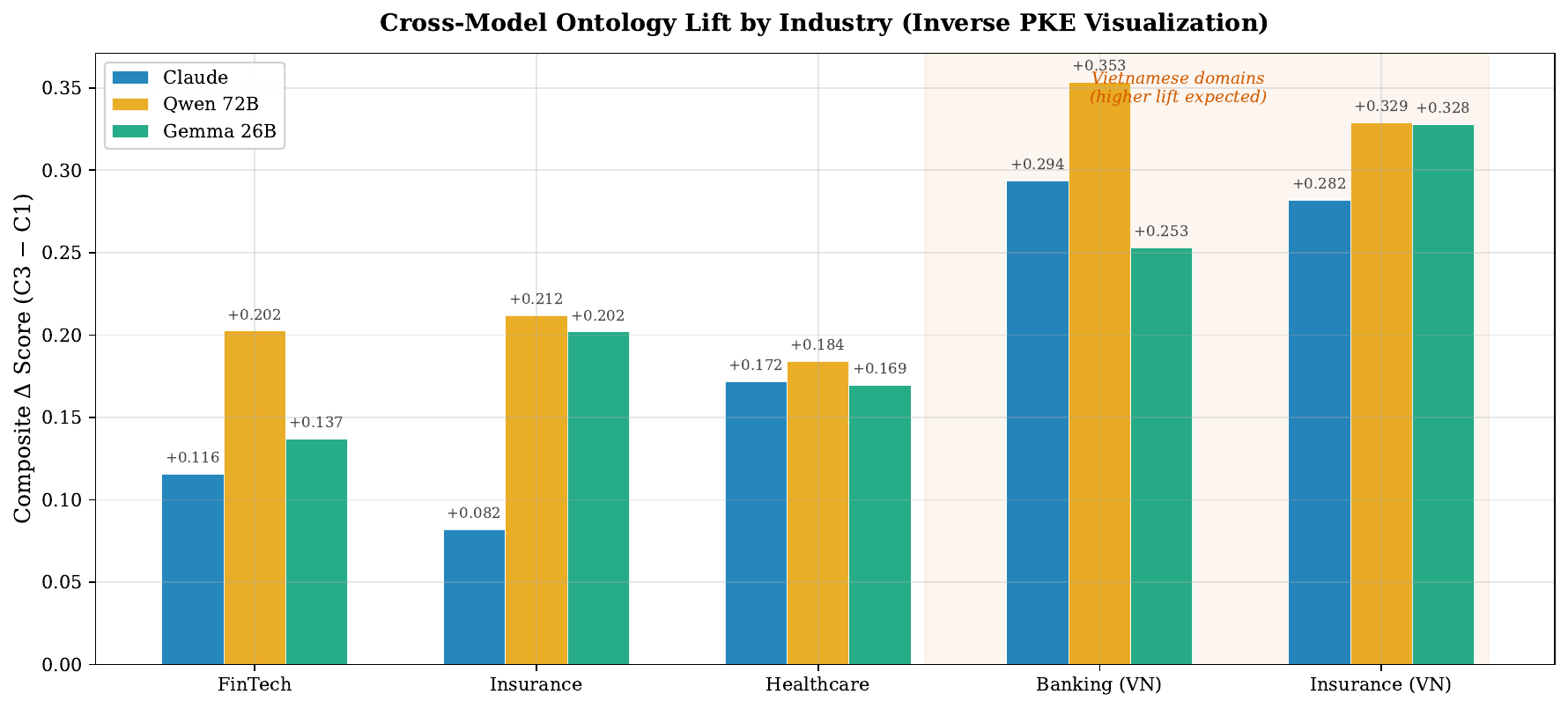}
\caption{Ontology lift ($\Delta_{\text{C1}\to\text{C3}}$) by industry
across three generator models. Vietnamese industries (shaded region)
consistently show larger improvement than English industries across all
models. Open-source models (Qwen, Gemma) benefit more than Claude,
supporting the Inverse PKE at both domain and model levels.}
\label{fig:crossmodel}
\end{figure}

The ontology lift replicates across models: Claude $+.15$, Qwen $+.22$,
Gemma $+.20$. A Wilcoxon signed-rank test found Qwen's lift
significantly larger than Claude's ($W = 370$, $p = .040$)---consistent
with Inverse PKE at the model level: the open-source model benefits
more from grounding because it has less parametric coverage. Vietnamese
amplification replicates in all three models (Vietnamese $\Delta$
exceeds English $\Delta$ by $1.5$--$2.0\times$;
Figure~\ref{fig:crossmodel}), indicating that Inverse PKE operates at both
domain and model levels.

\subsection{Threats to Validity}

\begin{enumerate}
  \item \textbf{Ontology completeness}: If a domain concept is missing,
    grounding is incomplete. Maintenance cost scales with regulatory
    change velocity.
  \item \textbf{OWL reasoning practicality}: Proposed output-side
    validation via description logic reasoners (\S\ref{sec:closed-loop})
    requires translating free-text LLM outputs into OWL-compatible
    representations---itself an error-prone process.
  \item \textbf{Single-system analysis}: Findings are based on one
    platform (\faos{}); cross-model replication (\Cref{sec:crossmodel})
    addresses model-dependence but replication on CrewAI, AutoGen, or
    standalone LangGraph is future work.
  \item \textbf{LLM-as-judge reliability}: We rely on an LLM judge
    (Claude Sonnet~4) rather than human experts.
    \citet{schroeder2024judge}, \citet{gu2025llmjudge}, and
    \citet{szymanski2025iui} document LLM-judge failure modes; the latter
    specifically report SME--LLM agreement of only $64$--$68\%$ on
    expert tasks. We mitigate with three repetitions, temperature 0.0,
    and structured JSON output, but have not yet validated inter-rater
    reliability ($\kappa$) against domain experts. As a sensitivity
    envelope: under a pessimistic $35\%$ disagreement rate, our MA
    grand-mean effect ($\Delta_{\text{C1}\to\text{C3}} = +.37$;
    C1$=.372$, C3$=.740$) would compress to approximately $+.24$ after
    distributional correction---still a large effect. The entropy
    signature in \S\ref{sec:inverse-pke} provides a second line of
    evidence independent of judge calibration.
  \item \textbf{Curated RAG baseline}: C2 used chunks extracted from
    the same ontology blueprints as C3, making C2 unusually
    well-structured. Production RAG with heterogeneous documents would
    likely show a larger ontology advantage, consistent with
    \citet{sharma2025ograg}'s $+55\%$ fact-recall result.
  \item \textbf{Context interference}: The TF regression suggests
    injection can displace parametric knowledge---independently
    validated by \citet{lin2026resisting} ($+22.89\%$ from resisting
    contextual interference) and \citet{du2025context} ($13.9$--$85\%$
    degradation from context length). Warrants investigation with
    adaptive injection strategies.
  \item \textbf{Ecological validity}: The ontology content is the same
    content deployed to production agents across 22 industry verticals,
    providing ecological validity beyond laboratory conditions.
\end{enumerate}

\section{Conclusion and Future Work}
\label{sec:conclusion}

We have presented a neurosymbolic architecture for enterprise AI agents
that uses a three-layer ontological framework (Role, Domain,
Interaction) to ground LLM reasoning in formal domain knowledge. Our
primary contribution is the formalization of \emph{asymmetric
neurosymbolic coupling}---the observation that current enterprise
systems constrain agent inputs but not outputs---and the proposal of a
\emph{closed-loop neurosymbolic framework} extending ontological
constraints to output validation and ontology evolution. Our framework
distinguishes \textbf{implemented} mechanisms (L1--L3: three-layer
ontology, context injection, tool discovery, governance thresholds,
autonomy gates, quality judge---deployed across 22 industry verticals
with $650$+ specialized agents) from \textbf{proposed} extensions
(L4--L5: output-side validation, OWL constraint checking, supervised
ontology evolution---formalized architecturally but not yet empirically
evaluated).

Our empirical evaluation across 1,800 runs (600 per model) spanning
five regulated industries and three LLM architectures (Claude
Sonnet~4, Qwen~2.5 72B, Gemma~4 26B) provides strong,
model-independent evidence for ontological grounding. Ontology-coupled
agents significantly outperformed ungrounded agents on Metric Accuracy
($p < .001$, $W = .460$) and Role Consistency ($p < .001$, $W = .614$)
across all three models; Regulatory Compliance was model-sensitive
(significant on Claude and Qwen, $p = .051$ on Gemma). The strongest
improvements appear where LLM parametric knowledge is weakest:
Vietnamese industries showed $2\times$ the improvement of
English-language domains ($\Delta = +.29$ vs.\ $+.12$), replicating
across all three models. This pattern supports the claim that
ontological grounding is operationally important for non-English
enterprise domains where parametric coverage is structurally
insufficient. The multi-tenant
nature of \faos{} introduces an additional neurosymbolic challenge
(\emph{ontological polymorphism}): the same three-layer schema must
support distinct domain instantiations per tenant---a direction for
continued architectural work.

Future work targets four areas:

\begin{enumerate}
  \item \textbf{Layer ablation}: factorial design isolating the
    contribution of each ontology layer ($\role$, $\domain$,
    $\interact$) to each metric.
  \item \textbf{Implementing L4 validation}: building and evaluating
    the \texttt{OntologyValidator}, focusing on latency--quality
    trade-offs; our companion work
    \citep{luong2026verification} provides the simulation and
    certification framework.
  \item \textbf{Adaptive context injection}: selective injection based
    on estimated parametric knowledge $\kappa(d)$ to address the TF
    regression on well-known concepts. Self-RAG
    \citep{asai2024selfrag} uses reflection tokens to decide
    \emph{when to retrieve}; analogous \emph{when to inject}
    mechanisms are formalized as an information-theoretic optimization
    in \citep{luong2026context}.
  \item \textbf{Full-scale evaluation}: human-expert validation
    ($\kappa$ inter-rater reliability), noisy RAG baselines, and
    additional industry verticals.
\end{enumerate}

Neurosymbolic architectures are not merely beneficial but
\emph{necessary} for enterprise AI agents in regulated industries.
The question is not whether to integrate symbolic and neural
reasoning, but how to do so with minimal latency, maximal coverage,
and provable guarantees.

\section*{Acknowledgments}

The FAOS platform, including ontology blueprints and experiment
infrastructure, was developed by the first author as part of the
Foundation AgenticOS project. This work originates from the doctoral
research program at Golden Gate University.

\section*{Data and Code Availability}

The experiment harness, the complete 50-task definition set
(10 per industry, covering Terminology, Metric, Regulatory, Role, and
cross-cutting categories) with ontology-derived ground truth, condition
implementations, LLM-as-judge rubrics, and aggregated results
(per-condition summaries, cross-model analyses, entropy analyses) are
publicly available at the FAOS Research repository
\citep{faos-research-repo}. All statistical analyses are performed
exclusively on the 600 planned runs per model
(50 tasks $\times$ 4 conditions $\times$ 3 repetitions); 386
supplementary rows from early development are preserved in the
repository archive for transparency but excluded from all reported
results. Ontology blueprints are drawn from the \faos{} platform's
production ontology definitions for FinTech, Insurance, Healthcare,
Vietnamese Banking, and Vietnamese Insurance verticals. Raw per-run
CSV outputs and full judge transcripts will be released via the same
repository upon paper acceptance.

\clearpage
{\footnotesize
\setlength{\bibsep}{0pt plus 0.2ex}
\bibliography{references}

@article{garcez2019neural,
  author    = {Garcez, Artur d'Avila and Lamb, Luis C.},
  title     = {Neurosymbolic {AI}: The 3rd Wave},
  journal   = {Artificial Intelligence Review},
  year      = {2023},
  volume    = {56},
  pages     = {12387--12406},
  doi       = {10.1007/s10462-023-10448-w},
  note      = {Originally circulated 2019; published 2023}
}

@incollection{hitzler2022neuro,
  author    = {Hitzler, Pascal and Sarker, Md Kamruzzaman},
  title     = {Neuro-Symbolic Artificial Intelligence: The State of the Art},
  booktitle = {Neuro-Symbolic Artificial Intelligence: The State of the Art},
  publisher = {IOS Press},
  year      = {2022},
  series    = {Frontiers in Artificial Intelligence and Applications},
  volume    = {342},
  doi       = {10.3233/FAIA342}
}

@article{fodor1988connectionism,
  author    = {Fodor, Jerry A. and Pylyshyn, Zenon W.},
  title     = {Connectionism and Cognitive Architecture: A Critical Analysis},
  journal   = {Cognition},
  year      = {1988},
  volume    = {28},
  number    = {1--2},
  pages     = {3--71},
  doi       = {10.1016/0010-0277(88)90031-5}
}

@article{ji2023hallucination,
  author    = {Ji, Ziwei and Lee, Nayeon and Frieske, Rita and Yu, Tiezheng
               and Su, Dan and Xu, Yan and Ishii, Etsuko and Bang, Yejin
               and Madotto, Andrea and Fung, Pascale},
  title     = {Survey of Hallucination in Natural Language Generation},
  journal   = {ACM Computing Surveys},
  year      = {2023},
  volume    = {55},
  number    = {12},
  pages     = {1--38},
  doi       = {10.1145/3571730}
}

@article{huang2023survey,
  author    = {Huang, Lei and Yu, Weijiang and Ma, Weitao and Zhong, Weihong
               and Feng, Zhangyin and Wang, Haotian and Chen, Qianglong
               and Peng, Weihua and Feng, Xiaocheng and Qin, Bing
               and Liu, Ting},
  title     = {A Survey on Hallucination in Large Language Models: Principles,
               Taxonomy, Challenges, and Open Questions},
  journal   = {arXiv preprint arXiv:2311.05232},
  year      = {2023}
}

@inproceedings{lewis2020retrieval,
  author    = {Lewis, Patrick and Perez, Ethan and Piktus, Aleksandra
               and Petroni, Fabio and Karpukhin, Vladimir and Goyal, Naman
               and K{\"u}ttler, Heinrich and Lewis, Mike and Yih, Wen-tau
               and Rockt{\"a}schel, Tim and Riedel, Sebastian
               and Kiela, Douwe},
  title     = {Retrieval-Augmented Generation for Knowledge-Intensive {NLP}
               Tasks},
  booktitle = {Advances in Neural Information Processing Systems (NeurIPS)},
  year      = {2020},
  volume    = {33},
  pages     = {9459--9474}
}

@article{gao2024retrieval,
  author    = {Gao, Yunfan and Xiong, Yun and Gao, Xinyu and Jia, Kangxiang
               and Pan, Jinliu and Bi, Yuxi and Dai, Yi and Sun, Jiawei
               and Wang, Haofen},
  title     = {Retrieval-Augmented Generation for Large Language Models:
               A Survey},
  journal   = {arXiv preprint arXiv:2312.10997},
  year      = {2024}
}

@article{pan2024unifying,
  author    = {Pan, Shirui and Luo, Linhao and Wang, Yufei and Chen, Chen
               and Wang, Jiapu and Wu, Xindong},
  title     = {Unifying Large Language Models and Knowledge Graphs:
               A Roadmap},
  journal   = {IEEE Transactions on Knowledge and Data Engineering},
  year      = {2024},
  volume    = {36},
  number    = {7},
  pages     = {3580--3599},
  doi       = {10.1109/TKDE.2024.3352100}
}

@inproceedings{zhang2022greaselm,
  author    = {Zhang, Xikun and Bosselut, Antoine and Yasunaga, Michihiro
               and Ren, Hongyu and Liang, Percy and Manning, Christopher D.
               and Leskovec, Jure},
  title     = {{GreaseLM}: Graph {REASoning} Enhanced Language Models},
  booktitle = {International Conference on Learning Representations (ICLR)},
  year      = {2022}
}

@article{babaei2023llms4ol,
  author    = {Babaei Giglou, Hamed and D'Souza, Jennifer and Auer, S{\"o}ren},
  title     = {{LLMs4OL}: Large Language Models for Ontology Learning},
  journal   = {arXiv preprint arXiv:2307.16648},
  year      = {2023}
}

@article{willard2023efficient,
  author    = {Willard, Brandon T. and Louf, R{\'e}mi},
  title     = {Efficient Guided Generation for Large Language Models},
  journal   = {arXiv preprint arXiv:2307.09702},
  year      = {2023}
}

@inproceedings{wei2022chain,
  author    = {Wei, Jason and Wang, Xuezhi and Schuurmans, Dale
               and Bosma, Maarten and Ichter, Brian and Xia, Fei
               and Chi, Ed H. and Le, Quoc V. and Zhou, Denny},
  title     = {Chain-of-Thought Prompting Elicits Reasoning in Large
               Language Models},
  booktitle = {Advances in Neural Information Processing Systems (NeurIPS)},
  year      = {2022},
  volume    = {35},
  pages     = {24824--24837}
}

@inproceedings{wang2023selfconsistency,
  author    = {Wang, Xuezhi and Wei, Jason and Schuurmans, Dale
               and Le, Quoc V. and Chi, Ed H. and Narang, Sharan
               and Chowdhery, Aakanksha and Zhou, Denny},
  title     = {Self-Consistency Improves Chain of Thought Reasoning in
               Language Models},
  booktitle = {International Conference on Learning Representations (ICLR)},
  year      = {2023}
}

@inproceedings{hu2022lora,
  author    = {Hu, Edward J. and Shen, Yelong and Wallis, Phillip
               and Allen-Zhu, Zeyuan and Li, Yuanzhi and Wang, Shean
               and Wang, Lu and Chen, Weizhu},
  title     = {{LoRA}: Low-Rank Adaptation of Large Language Models},
  booktitle = {International Conference on Learning Representations (ICLR)},
  year      = {2022}
}

@article{fox1992enterprise,
  author    = {Fox, Mark S. and Barbuceanu, Mihai and Gruninger, Michael},
  title     = {An Organisation Ontology for Enterprise Modelling:
               Preliminary Concepts for Linking Structure and Behaviour},
  journal   = {Computers in Industry},
  year      = {1996},
  volume    = {29},
  number    = {1--2},
  pages     = {123--134},
  doi       = {10.1016/0166-3615(95)00079-8},
  note      = {Originally presented 1992}
}

@article{bennett2013financial,
  author    = {Bennett, Mike},
  title     = {The Financial Industry Business Ontology: Best Practice
               for Big Data},
  journal   = {Journal of Banking Regulation},
  year      = {2013},
  volume    = {14},
  number    = {3--4},
  pages     = {255--268},
  doi       = {10.1057/jbr.2013.13}
}

@book{arp2015building,
  author    = {Arp, Robert and Smith, Barry and Spear, Andrew D.},
  title     = {Building Ontologies with {Basic Formal Ontology}},
  publisher = {MIT Press},
  year      = {2015},
  isbn      = {978-0-262-52781-5}
}

@article{borgo2022dolce,
  author    = {Borgo, Stefano and Ferrario, Roberta and Gangemi, Aldo
               and Guarino, Nicola and Masolo, Claudio and Porello, Daniele
               and Sanfilippo, Emilio M. and Vieu, Laure},
  title     = {{DOLCE}: A Descriptive Ontology for Linguistic and
               Cognitive Engineering},
  journal   = {Applied Ontology},
  year      = {2022},
  volume    = {17},
  pages     = {45--69},
  doi       = {10.3233/AO-210259}
}

@inproceedings{guo2024largelanguage,
  author    = {Guo, Taicheng and Chen, Xiuying and Wang, Yaqi and
               Chang, Ruidi and Pei, Shichao and Chawla, Nitesh V. and
               Wiest, Olaf and Zhang, Xiangliang},
  title     = {Large Language Model Based Multi-Agents: A Survey of
               Progress and Challenges},
  booktitle = {Proceedings of the 33rd International Joint Conference on
               Artificial Intelligence ({IJCAI})},
  year      = {2024},
  pages     = {8048--8057},
  url       = {https://arxiv.org/abs/2402.01680},
  note      = {Survey consolidating classical MAS formalisms and
               LLM-agent behavior, with explicit treatment of how
               LLM stochasticity breaks well-defined utility/protocol
               assumptions}
}

@book{evans2004domain,
  author    = {Evans, Eric},
  title     = {Domain-Driven Design: Tackling Complexity in the Heart
               of Software},
  publisher = {Addison-Wesley},
  year      = {2004},
  isbn      = {978-0-321-12521-7}
}

@article{harnad1990symbol,
  author    = {Harnad, Stevan},
  title     = {The Symbol Grounding Problem},
  journal   = {Physica D: Nonlinear Phenomena},
  year      = {1990},
  volume    = {42},
  number    = {1--3},
  pages     = {335--346},
  doi       = {10.1016/0167-2789(90)90087-6}
}

@article{mialon2023augmented,
  author    = {Mialon, Gr{\'e}goire and Dess{\`i}, Roberto and Lomeli, Maria
               and Nalmpantis, Christoforos and Pasunuru, Ram and Raber, Roberta
               and Roziere, Baptiste and Schick, Timo and Dwivedi-Yu, Jane
               and Celikyilmaz, Asli and Grave, Edouard and LeCun, Yann
               and Scialom, Thomas},
  title     = {Augmented Language Models: A Survey},
  journal   = {Transactions on Machine Learning Research (TMLR)},
  year      = {2023}
}

@inproceedings{kambhampati2024llms,
  author    = {Kambhampati, Subbarao and Valmeekam, Karthik and Guan, Lin
               and Verma, Mudit and Stechly, Kaya and Bhambri, Siddhant
               and Saldyt, Lucas and Murthy, Anil},
  title     = {Position: {LLMs} Can't Plan, But Can Help Planning in
               {LLM}-Modulo Frameworks},
  booktitle = {Proceedings of the 41st International Conference on Machine
               Learning (ICML)},
  year      = {2024},
  note      = {arXiv:2402.01817}
}

@article{marcus2020next,
  author    = {Marcus, Gary},
  title     = {The Next Decade in {AI}: Four Steps Towards Robust
               Artificial Intelligence},
  journal   = {arXiv preprint arXiv:2002.06177},
  year      = {2020}
}

@misc{euaiact2024,
  author    = {{European Parliament and Council}},
  title     = {Regulation ({EU}) 2024/1689 --- Artificial Intelligence Act},
  year      = {2024},
  howpublished = {Official Journal of the European Union, L series},
  note      = {Entered into force August 1, 2024}
}

@article{venkatesh2026ontollm,
  author    = {Venkatesh, Pruthvi Raj and Radha Krishna, P.},
  title     = {{OntoLLM}: Enhancing {LLM} Grounding and Digression
               Prevention with Ontologies and Knowledge Graphs},
  journal   = {Expert Systems with Applications},
  year      = {2026},
  doi       = {10.1016/j.eswa.2026.126418},
  note      = {ScienceDirect S0957417426004185; published February 2026}
}

@article{dacruz2025ontology,
  author    = {da Cruz, Tiago and Asprino, Luigi and Presutti, Valentina},
  title     = {Ontology Learning and Knowledge Graph Construction:
               A Comparison of Approaches and Their Impact on {RAG}
               Performance},
  journal   = {arXiv preprint arXiv:2511.05991},
  year      = {2025},
  note      = {Submitted November 2025; compares ontology-derived,
               text-derived, and vector-only retrieval for RAG performance}
}

@inproceedings{agrawal2024can,
  author    = {Agrawal, Garima and Kumarage, Tharindu and
               Alghamdi, Zeyad and Liu, Huan},
  title     = {Can Knowledge Graphs Reduce Hallucinations in {LLMs}?
               {A} Survey},
  booktitle = {Proceedings of the 2024 Conference of the North American
               Chapter of the Association for Computational Linguistics
               (NAACL)},
  year      = {2024},
  pages     = {3947--3960},
  url       = {https://aclanthology.org/2024.naacl-long.219/}
}

@article{sansford2024grapheval,
  author    = {Sansford, Hannah and Richardson, Nicholas and
               Petric Maretic, Hermina and Nait Saada, Juba},
  title     = {{GraphEval}: A Knowledge-Graph Based {LLM} Hallucination
               Evaluation Framework},
  journal   = {arXiv preprint arXiv:2407.10793},
  year      = {2024}
}

@inproceedings{zheng2023judging,
  author    = {Zheng, Lianmin and Chiang, Wei-Lin and Sheng, Ying
               and Zhuang, Siyuan and Wu, Zhanghao and Zhuang, Yonghao
               and Lin, Zi and Li, Zhuohan and Li, Dacheng and Xing, Eric P.
               and Zhang, Hao and Gonzalez, Joseph E. and Stoica, Ion},
  title     = {Judging {LLM}-as-a-Judge with {MT-Bench} and {Chatbot Arena}},
  booktitle = {Advances in Neural Information Processing Systems
               (NeurIPS), Datasets and Benchmarks Track},
  year      = {2023},
  volume    = {36},
  note      = {arXiv:2306.05685}
}

@article{friedman1937use,
  author    = {Friedman, Milton},
  title     = {The Use of Ranks to Avoid the Assumption of Normality
               Implicit in the Analysis of Variance},
  journal   = {Journal of the American Statistical Association},
  year      = {1937},
  volume    = {32},
  number    = {200},
  pages     = {675--701},
  doi       = {10.1080/01621459.1937.10503522}
}

@inproceedings{sharma2025ograg,
  author    = {Sharma, Kartik and Kumar, Peeyush and Li, Yunqing},
  title     = {{OG-RAG}: Ontology-Grounded Retrieval-Augmented Generation
               for Large Language Models},
  booktitle = {Proceedings of the 2025 Conference on Empirical Methods in
               Natural Language Processing (EMNLP)},
  year      = {2025},
  note      = {+55\% fact recall, +40\% response correctness via
               ontology-anchored hypergraph retrieval across 4 LLMs}
}

@article{ali2026clinical,
  author    = {Ali, Mohamed and Taha, Zaki and Morsey, Mohamed Mabrouk},
  title     = {Ontology-Grounded Knowledge Graphs for Mitigating Hallucinations
               in Large Language Models for Clinical Question Answering},
  journal   = {Journal of Biomedical Informatics},
  year      = {2026},
  note      = {98\% accuracy vs.\ 37\% for ChatGPT-4; hallucination
               reduced from 63\% to 1.7\% via RDF/OWL ontology + KG}
}

@inproceedings{yang2025nesyai,
  author    = {Yang, Xiao-Wen and Shao, Jie-Jing and Guo, Lan-Zhe
               and Zhang, Bo-Wen and Zhou, Zhi and Jia, Lin-Han
               and Dai, Wang-Zhou and Li, Yu-Feng},
  title     = {Neuro-Symbolic Artificial Intelligence: Towards Improving
               the Reasoning Abilities of Large Language Models},
  booktitle = {Proceedings of the Thirty-Fourth International Joint Conference
               on Artificial Intelligence (IJCAI), Survey Track},
  year      = {2025},
  note      = {First comprehensive IJCAI survey; Symbolic$\to$LLM,
               LLM$\to$Symbolic, LLM+Symbolic taxonomy}
}

@article{amayuelas2025grounding,
  author    = {Amayuelas, Alfonso and Sain, Joy and Kaur, Simerjot
               and Smiley, Charese},
  title     = {Grounding {LLM} Reasoning with Knowledge Graphs},
  journal   = {arXiv preprint arXiv:2502.13247},
  year      = {2025},
  note      = {+26.5\% over CoT baselines on GRBench; links each
               reasoning step to graph-structured data}
}

@article{schroeder2024judge,
  author    = {Schroeder, Kayla and Wood-Doughty, Zach},
  title     = {Can You Trust {LLM} Judgments? Reliability of
               {LLM}-as-a-Judge},
  journal   = {arXiv preprint arXiv:2412.12509},
  year      = {2024},
  note      = {Submitted December 2024; demonstrates limitations of
               single-shot LLM evaluation via McDonald's omega and
               internal-consistency reliability}
}

@inproceedings{lin2026resisting,
  author    = {Lin, Chenyu and Wen, Yilin and Su, Du and Tan, Hexiang
               and Sun, Fei and Chen, Muhan and Bao, Chenfu and Lyu, Zhonghou},
  title     = {Resisting Contextual Interference in {RAG} via
               Parametric-Knowledge Reinforcement},
  booktitle = {Proceedings of the International Conference on Learning
               Representations (ICLR)},
  year      = {2026},
  note      = {Knowledgeable-R1: +22.89\% in counterfactual scenarios
               by resisting parametric displacement}
}

@inproceedings{du2025context,
  author    = {Du, Yufeng and Tian, Minyang and Ronanki, Srikanth and others},
  title     = {Context Length Alone Hurts {LLM} Performance Despite
               Perfect Retrieval},
  booktitle = {Findings of the Association for Computational Linguistics:
               EMNLP 2025},
  year      = {2025},
  note      = {Performance degrades 13.9--85\% from length alone}
}

@article{tang2025parametric,
  author    = {Tang, Minghao and Ni, Shiyu and Wu, Jingtong and Han, Zengxin
               and Bi, Keping},
  title     = {Understanding Parametric Knowledge Injection in
               Retrieval-Augmented Generation},
  journal   = {arXiv preprint arXiv:2510.12668},
  year      = {2025},
  note      = {P-RAG does not consistently outperform T-RAG;
               hybrid PT-RAG achieves best performance}
}

@article{luong2026context,
  author    = {Luong Tuan, Thanh},
  title     = {The Information-Theoretic Bounds of Ontological Context:
               Optimal Knowledge Injection for Enterprise {LLM} Agents},
  journal   = {arXiv preprint},
  year      = {2026},
  note      = {RA-4 companion paper; 2,400-run experiment planned across
               5 industries and 16 context volume$\times$structure conditions}
}

@article{luong2026verification,
  author    = {Luong Tuan, Thanh and Sanyal, Abhijit},
  title     = {Toward Verifiable Enterprise {AI} Agents: Ontology-Powered
               Simulation and Formal Safety Certification},
  journal   = {arXiv preprint},
  year      = {2026},
  note      = {RA-6 companion paper; 1,800 scenarios across 5 industries,
               ontology-powered verification framework}
}

@inproceedings{liu2025ontotune,
  author    = {Liu, Zhiqiang and Gan, Chengtao and Wang, Junjie and
               Zhang, Yichi and Bo, Zhongpu and Sun, Mengshu and
               Chen, Huajun and Zhang, Wen},
  title     = {{OntoTune}: Ontology-Driven Self-training for Aligning
               Large Language Models},
  booktitle = {Proceedings of the ACM Web Conference 2025 (WWW)},
  year      = {2025},
  doi       = {10.1145/3696410.3714816},
  note      = {arXiv:2502.05478; SNOMED CT ontology-driven LLM alignment
               via in-context learning self-training; code at
               github.com/zjukg/OntoTune}
}

@inproceedings{luo2025gcr,
  author    = {Luo, Linhao and Zhao, Zicheng and Haffari, Reza and
               Li, Yuan-Fang and Gong, Chen and Pan, Shirui},
  title     = {Graph-Constrained Reasoning: Faithful Reasoning on
               Knowledge Graphs with Large Language Models},
  booktitle = {Proceedings of the 42nd International Conference on
               Machine Learning (ICML)},
  year      = {2025},
  pages     = {41540--41565},
  note      = {arXiv:2410.13080; PMLR 267:41540--41565; KG-Trie constrains
               LLM decoding to valid KG paths; zero-shot generalization
               to unseen KGs}
}

@inproceedings{xu2024knowledge,
  author    = {Xu, Rongwu and Qi, Zehan and Guo, Zhijiang and
               Wang, Cunxiang and Wang, Hongru and Zhang, Yue and
               Xu, Wei},
  title     = {Knowledge Conflicts for {LLMs}: A Survey},
  booktitle = {Proceedings of the 2024 Conference on Empirical Methods in
               Natural Language Processing (EMNLP)},
  year      = {2024},
  pages     = {8541--8565},
  url       = {https://aclanthology.org/2024.emnlp-main.486/},
  note      = {arXiv:2403.08319; taxonomizes context-memory, inter-context,
               and intra-memory knowledge conflicts in LLMs}
}

@article{colelough2025nesyai,
  author    = {Colelough, Brandon C. and Regli, William},
  title     = {Neuro-Symbolic {AI} in 2024: A Systematic Review},
  journal   = {arXiv preprint arXiv:2501.05435},
  year      = {2025},
  note      = {PRISMA systematic review of 167 NeSyAI papers (from 1{,}428
               screened) across learning, inference, and knowledge
               representation; published January 2025}
}

@inproceedings{zhou2025metagent,
  author    = {Zhou, Yanfang and Liu, Yuntao and Li, Xiaodong and
               Zhao, Yongqiang and Wang, Xintong and Tian, Jinlong and
               Li, Zhenyu and Xu, Xinhai},
  title     = {{Metagent-P}: A Neuro-Symbolic Planning Agent with
               Metacognition for Open Worlds},
  booktitle = {Findings of the Association for Computational Linguistics:
               ACL 2025},
  year      = {2025},
  pages     = {22747--22764},
  url       = {https://aclanthology.org/2025.findings-acl.1169/},
  note      = {Symbolic planning verification + LLM world knowledge;
               reduces replanning by 34\% and exceeds the average human
               success rate by 18.96\% in long-horizon open-world tasks}
}

@article{peer2025ata,
  author    = {Peer, David and Stabinger, Sebastian},
  title     = {{ATA}: A Neuro-Symbolic Approach to Implement Autonomous
               and Trustworthy Agents},
  journal   = {arXiv preprint arXiv:2510.16381},
  year      = {2025},
  note      = {Submitted October 2025; decouples LLM agents into offline
               symbolic knowledge base ingestion and online deterministic
               execution}
}

@article{luong2026quantum,
  author    = {Luong Tuan, Thanh},
  title     = {Knowledge Interference in Enterprise {LLM} Agents:
               A Quantum-Inspired Framework for Context Engineering},
  journal   = {arXiv preprint},
  year      = {2026},
  note      = {RA-11 companion paper; formalizes constructive/destructive
               knowledge interference using quantum-inspired mathematics,
               with empirical validation from RA-3 and RA-4 datasets}
}

@article{gaurav2025governance,
  author    = {Gaurav, Suyash and Heikkonen, Jukka and
               Chaudhary, Jatin and Pervez, Helen},
  title     = {Governance-as-a-Service: A Multi-Agent Framework for
               {AI} System Compliance and Policy Enforcement},
  journal   = {arXiv preprint arXiv:2508.18765},
  year      = {2025},
  note      = {Governance-as-a-Service with Trust Factor scoring for
               multi-agent compliance enforcement; submitted Aug 2025}
}

@inproceedings{shen2025slot,
  author    = {Shen, Zhengyuan and Wang, Darren Yow-Bang and
               Mishra, Soumya Smruti and Xu, Zhichao and
               Teng, Yifei and Ding, Haibo},
  title     = {{SLOT}: Structuring the Output of Large Language Models},
  booktitle = {Proceedings of the 2025 Conference on Empirical Methods
               in Natural Language Processing: Industry Track (EMNLP)},
  year      = {2025},
  url       = {https://aclanthology.org/2025.emnlp-industry.32/},
  note      = {arXiv:2505.04016; model-agnostic post-processing achieving
               99.5\% schema accuracy for LLM output structuring}
}

@article{augenstein2026interplay,
  author    = {Augenstein, Isabelle},
  title     = {Understanding the Interplay between {LLMs}' Utilisation
               of Parametric and Contextual Knowledge: A Keynote at
               {ECIR} 2025},
  journal   = {arXiv preprint arXiv:2603.09654},
  year      = {2026},
  note      = {ECIR 2025 keynote; characterises how parametric--contextual
               interaction shapes LLM behaviour; strengthens Inverse PKE
               theoretical grounding}
}

@article{qwen2025qwen25,
  author    = {Yang, An and Yang, Baosong and Zhang, Beichen and Hui, Binyuan and
               Wang, Bo and Zheng, Bowen and Yu, Chengyuan and others},
  title     = {Qwen2.5 Technical Report},
  journal   = {arXiv preprint arXiv:2412.15115},
  year      = {2025},
  note      = {Qwen Team, Alibaba Group}
}

@article{farquhar2024semantic,
  author    = {Farquhar, Sebastian and Kossen, Jannik and Kuhn, Lorenz and Gal, Yarin},
  title     = {Detecting Hallucinations in Large Language Models Using Semantic Entropy},
  journal   = {Nature},
  volume    = {630},
  pages     = {625--630},
  year      = {2024},
  doi       = {10.1038/s41586-024-07421-0}
}

@inproceedings{kossen2025sep,
  author    = {Kossen, Jannik and Han, Jiatong and Razzak, Muhammed and Schut, Lisa and Malik, Shreshth and Gal, Yarin},
  title     = {Semantic Entropy Probes: Robust and Cheap Hallucination Detection in {LLMs}},
  booktitle = {Proceedings of the International Conference on Learning Representations (ICLR)},
  year      = {2025},
  note      = {arXiv:2406.15927}
}

@article{blondel2025arms,
  author    = {Blondel, Mathieu and Sander, Michael E. and Vivier-Ardisson, Gabriel and Liu, Taiji and Roulet, Vincent},
  title     = {Autoregressive Language Models are Secretly Energy-Based Models},
  journal   = {arXiv preprint arXiv:2512.15605},
  year      = {2025}
}

@article{liu2025neural,
  author    = {Liu, Ziming and Liu, Yizhou and Gore, Jeremy and Tegmark, Max},
  title     = {Neural Thermodynamic Laws for Large Language Model Training},
  journal   = {arXiv preprint arXiv:2505.10559},
  year      = {2025}
}

@article{wei2025senator,
  author    = {Wei, Yifan and Yu, Xiaoyan and Pan, Tengfei and
               Li, Angsheng and Du, Li},
  title     = {Structural Entropy Guided Agent for Detecting and
               Repairing Knowledge Deficiencies in {LLMs}},
  journal   = {arXiv preprint arXiv:2505.07184},
  year      = {2025},
  note      = {SENATOR: structural-entropy-guided knowledge navigator;
               improves Llama-3-8B by 11.98\% and Qwen2-7B by 9.15\%
               on medical benchmarks}
}

@misc{anthropic2025claude,
  author    = {{Anthropic}},
  title     = {Claude {Sonnet} 4 Model Card},
  year      = {2025},
  howpublished = {Anthropic Technical Documentation},
  note      = {Available at \texttt{docs.anthropic.com}. Accessed March 2026}
}

@article{gu2025llmjudge,
  author    = {Gu, Jiawei and Jiang, Xuhui and Shi, Zhichao and
               Tan, Hexiang and Zhai, Xuehao and Xu, Chengjin and
               Li, Wei and Shen, Yinghan and Ma, Shengjie and
               Liu, Honghao and Wang, Saizhuo and Zhang, Kun and
               Wang, Yuanzhuo and Gao, Wen and Ni, Lionel and Guo, Jian},
  title     = {A Survey on {LLM}-as-a-Judge},
  journal   = {arXiv preprint arXiv:2411.15594},
  year      = {2024},
  note      = {Submitted November 2024, last revised October 2025;
               comprehensive survey of LLM-as-judge failure modes:
               position bias, length bias, transitivity failures}
}

@inproceedings{szymanski2025iui,
  author    = {Szymanski, Annalisa and Ziems, Noah and
               Eicher-Miller, Heather A. and Li, Toby Jia-Jun and
               Jiang, Meng and Metoyer, Ronald A.},
  title     = {Limitations of the {LLM}-as-a-Judge Approach for Evaluating
               {LLM} Outputs in Expert Knowledge Tasks},
  booktitle = {Proceedings of the 30th International Conference on
               Intelligent User Interfaces (IUI)},
  year      = {2025},
  doi       = {10.1145/3708359.3712091},
  note      = {arXiv:2410.20266; SME--LLM judge agreement is only 64--68\%
               in dietetics + mental-health expert tasks}
}

@article{hakim2026nesyagentic,
  author    = {Hakim, Safayat Bin and Adil, Muhammad and
               Velasquez, Alvaro and Song, Houbing Herbert},
  title     = {Neuro-Symbolic Agentic {AI}: Architectures, Integration
               Patterns, Applications, Open Challenges and Future
               Research Directions},
  journal   = {Information Fusion},
  year      = {2026},
  doi       = {10.1016/j.inffus.2026.103110},
  note      = {ScienceDirect S1574013726000110; systematic PRISMA review
               of 178 neuro-symbolic agentic papers (2020--Nov 2025) with
               taxonomy of architectural configurations and integration
               dimensions: knowledge representation (44\%), learning and
               inference (63\%), logic and reasoning (35\%),
               explainability/trustworthiness (28\%), meta-cognition (5\%)}
}

@misc{skan2026aow,
  author    = {{Skan AI}},
  title     = {Agentic Ontology of Work ({AOW}): A Common Language for
               the Age of Intelligent Automation},
  year      = {2026},
  howpublished = {Industry White Paper},
  note      = {Open enterprise ontology with 8 entity types (Agents, Skills,
               Intents, Contexts, Policies, Memory, Confidence, Outcomes)}
}

@inproceedings{zhao2025reconciliation,
  author    = {Zhao, Jun and Yang, Yongzhuo and Hu, Xiang and
               Tong, Jingqi and Lu, Yi and Wu, Wei and Gui, Tao and
               Zhang, Qi and Huang, Xuanjing},
  title     = {Understanding Parametric and Contextual Knowledge
               Reconciliation within Large Language Models},
  booktitle = {Advances in Neural Information Processing Systems (NeurIPS)},
  year      = {2025},
  url       = {https://openreview.net/forum?id=76cFMRgEzQ},
  note      = {NeurIPS 2025 spotlight; models knowledge flow as entity flow
               and traces how parametric and contextual knowledge are
               reconciled through distinct attention heads with layer-wise
               accumulation}
}

@inproceedings{asai2024selfrag,
  author    = {Asai, Akari and Wu, Zeqiu and Wang, Yizhong and
               Sil, Avirup and Hajishirzi, Hannaneh},
  title     = {Self-{RAG}: Learning to Retrieve, Generate, and Critique
               through Self-Reflection},
  booktitle = {International Conference on Learning Representations (ICLR)},
  year      = {2024},
  note      = {Selective retrieval via reflection tokens; adaptive
               decision of when to retrieve}
}

@misc{faos-research-repo,
  author       = {Luong, Thanh Tuan and Sanyal, Abhijit},
  title        = {{FAOS}~Research: Code, Data, and Ontologies for Ontology-Powered Enterprise Agent Verification},
  year         = {2026},
  howpublished = {\url{https://github.com/frank-luongt/faos-research}},
  note         = {GitHub repository. Experiment harness, 50-task definitions, ontology context files, LLM-as-judge rubrics, and aggregated results. Raw per-run outputs released upon acceptance.}
}
}

\end{document}